\documentclass{article} 
\usepackage{iclr2027_conference,times}

\usepackage{amsmath,amsfonts,bm}

\def\eqref#1{equation~\ref{#1}}

\def\1{\bm{1}}

\DeclareMathAlphabet{\mathsfit}{\encodingdefault}{\sfdefault}{m}{sl}
\SetMathAlphabet{\mathsfit}{bold}{\encodingdefault}{\sfdefault}{bx}{n}

\definecolor{dblue}{rgb}{0, 0, 0.6}
\definecolor{dgreen}{rgb}{0,0.4,0.2}
\definecolor{dred}{rgb}{0.6, 0, 0}
\usepackage[colorlinks,
            linkcolor=magenta,
            anchorcolor=magenta,
            urlcolor=dgreen,
            citecolor=blue
            ]{hyperref}
\usepackage{url}

\usepackage{graphicx}
\usepackage{multirow}
\usepackage[normalem]{ulem}
\useunder{\uline}{\ul}{}
\usepackage{booktabs}
\usepackage{algorithm}
\usepackage{algpseudocode}
\usepackage{subcaption}
\usepackage{cleveref}
\usepackage{wrapfig}
\usepackage{subcaption}
\usepackage{enumitem}
\usepackage{needspace}

\usepackage{amsthm}

\theoremstyle{corollary}

\usepackage{makecell}

\usepackage{pifont}
\usepackage[table]{xcolor}

\newcommand{\cmark}{\textcolor{green!60!black}{\ding{51}}}
\newcommand{\xmark}{\textcolor{red!70!black}{\ding{55}}}

\usepackage{etoc}

\title{When Less Data Favors Smaller Teachers: Rethinking Teacher Capacity and Data Selection for Knowledge Distillation}

\author{
Minjae Park, Taesun Yeom \& Jaeho Lee \\
Pohang University of Science and Technology (POSTECH) \\
\texttt{\{mjae.park,tsyeom,jaeho.lee\}@postech.ac.kr}
}

\begin{document}

\etocdepthtag.toc{main}

\maketitle

\begin{abstract}

Data pruning reduces the training cost of knowledge distillation (KD). However, the preferred teacher capacity changes with the data budget: smaller teachers can outperform larger ones when limited training data are available. Understanding what drives this shift is important not only for teacher choice but also for identifying which samples are useful for distillation. We analyze teacher supervision by decomposing it into \textit{relational ordering}---the ranking of classes---and \textit{score geometry}---the magnitudes and margins of class probabilities---and show that the small-teacher advantage in the low-data regime arises not only from score geometry but also from relational ordering. Beyond understanding teacher capacity, our analysis reveals two properties of effective subsets: samples should match the difficulty appropriate for the available budget, and their relational signals should be diverse rather than redundant. Based on these findings, we propose \textbf{DVA} (\textbf{D}ifficulty- and \textbf{V}olume-\textbf{A}ware data selection for KD), a training-dynamics-free method, which uses a small teacher as a proxy for budget-aware difficulty filtering and class-conditional relational volume maximization. Despite requiring no training dynamics statistics, our method remains competitive with training-dynamics-based methods while consistently outperforming training-dynamics-free baselines.

\end{abstract}

\section{Introduction} {

Knowledge distillation (KD) \citep{hinton2015distilling} is a widely used approach for improving the performance of a smaller model (i.e., the student) by transferring knowledge from a larger model (i.e., the teacher). In recent years, KD and its variants have become increasingly important for compressing large-scale models, including large language models \citep{xu2024survey, team2024gemma, guo2025deepseek, yang2025qwen3} and vision foundation models \citep{zhang2023faster, simeoni2025dinov3}, into smaller models. 

However, training with KD on a large-scale dataset requires costly inference from the teacher model, substantially increasing the overall training cost. To address this issue, recent studies on KD have explored \textit{data pruning}---the selection of a smaller subset of training data---as a means of reducing training cost while preserving student performance \citep{chen2025medium, wu2026distill}. 

Beyond its computational benefits, however, data pruning can also alter which teacher is most effective for a given student. For example, \citet{benbaruch2025distilling} empirically observed that the optimal teacher varies with the available data budget (e.g., different data pruning ratios), with smaller teachers outperforming larger ones in the low-data regime. This observation reveals the limitation of the prevailing view that teacher quality is determined primarily by the capacity gap relative to the student  \citep{cho2019efficacy, mirzadeh2020improved}, pointing to an additional dimension that must be considered when selecting teachers for KD with data pruning.

\textbf{Contribution.} In this work, we  investigate the origin of this phenomenon and explore its broader implications. Specifically, we seek to answer the following research questions:
 \begin{itemize}[leftmargin=*,topsep=0pt,parsep=0pt,label={}]
    \item \textbf{Q1.} Why does the preferred teacher capacity change with the data budget? \hfill (\Cref{sec:2_obs}) 
    \item \textbf{Q2.} What should a pruned subset preserve under different data budgets in KD? \hfill(\Cref{sec:3_data_sel}) 
    \item \textbf{Q3.} How can these findings be turned into a practical data selection method?\hfill (\Cref{sec:4_method}) 
\end{itemize}

For \textbf{Q1}, we begin by examining how the teacher supervision itself changes with the model capacity. The larger teacher produces concentrated predictions on the target class, whereas the smaller teacher assigns  more probability to semantically related classes. Building on this observation, we decompose the teacher's predictions into two components to identify what drives this budget-dependent shift in teacher preference: \textit{Relational ordering}, which captures the ranking of classes, and \textit{score geometry}, which captures the magnitudes of class probabilities and the margins between them (\Cref{fig:overview_b}). In \Cref{sec:2_obs}, we show that score geometry---identified as an important factor in prior work \citep{li2022asymmetric,zhao2022decoupled}---is not the only factor underlying the small teacher's benefit in the low-data regime; relational ordering also has a substantial effect on performance.

Next, we move on to \textbf{Q2}. Here, our analyses reveal two properties that characterize high-performing pruned subsets at the sample and set levels. At the sample level, the preferred sample difficulty shifts toward easier samples as the data budget decreases. At the set level, among samples of comparable difficulty, subsets that preserve diverse relational information---i.e., the relative relationships between target and non-target predictions---outperform those dominated by redundant teacher outputs.

Based on these findings, for \textbf{Q3}, we propose \textbf{DVA} (\textbf{D}ifficulty- and \textbf{V}olume-\textbf{A}ware data selection), a training-dynamics-free framework that requires no statistics from student training runs. At the sample level, DVA filters examples based on budget-dependent difficulty. At the set level, DVA selects, within each class, a subset whose diverse relational signals collectively span a large volume in the teacher's output space. Empirically, across ResNet \citep{he2016deep} and vision transformer (ViT) \citep{dosovitskiy2021image} families on CIFAR-100, Tiny-ImageNet, and ImageNet, DVA consistently outperforms training-dynamics-free baselines while remaining competitive with training-dynamics-based methods, surpassing them in several settings.

Taken together, our study takes a step toward data budget-adaptive training methods for KD by investigating how the teacher's predictions should be shaped across different data budgets. We hope that our work contributes to a deeper understanding of the underlying mechanisms of KD.

\begin{figure*}[t]
    \centering
    \includegraphics[
        width=\textwidth,
        trim={0.7cm 0.3cm 0.5cm 0cm},
        clip
    ]{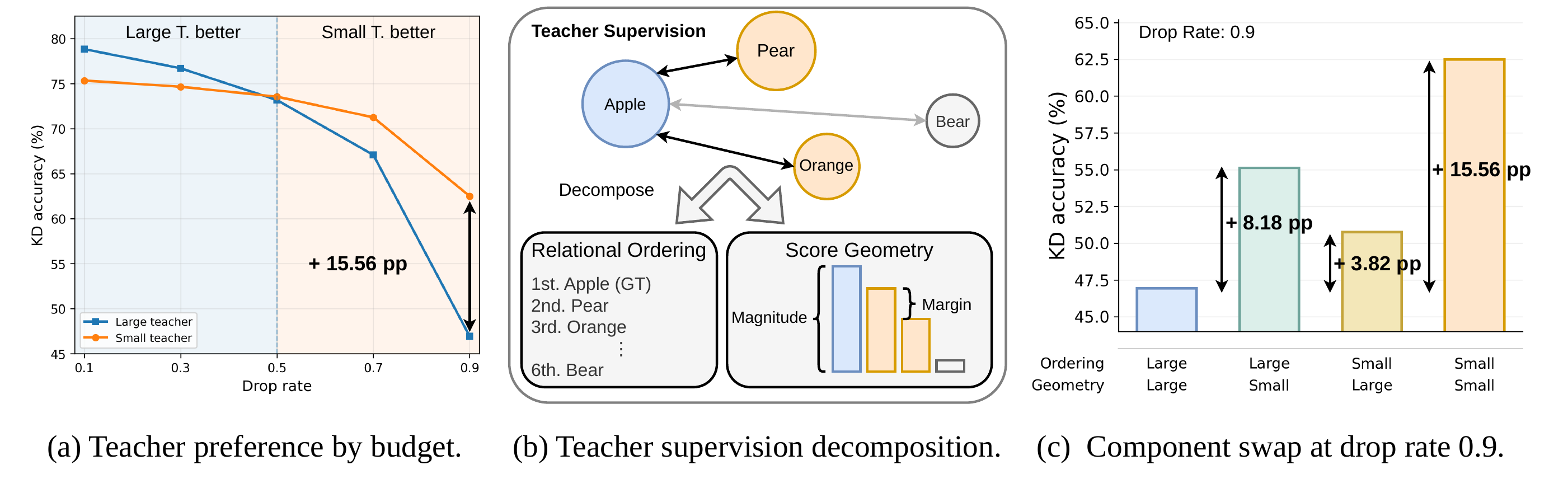}
    \caption{\textbf{Overview of our analysis.} Large and small denote ResNet-34 teachers with width 64 and 8; GT denotes the ground-truth class.
    (\textbf{a}) Under random pruning on CIFAR-100 with a ResNet-18 student, the preferred teacher capacity changes with the data budget. The preference reverses around a drop rate of 0.5, and at 0.9, the small teacher outperforms the large teacher by 15.56 percentage points.
    (\textbf{b}) We decompose teacher supervision into \textit{relational ordering} and \textit{score geometry}.
    (\textbf{c}) Replacing only its score geometry with that of the small teacher improves KD accuracy by 8.18 points, while replacing only the large teacher's relational ordering improves accuracy by 3.82 points, showing that both components contribute to the small-teacher advantage.
    }
    \label{fig:overview}
    \phantomsubcaption\label{fig:overview_a}
    \phantomsubcaption\label{fig:overview_b}
    \phantomsubcaption\label{fig:overview_c}
\end{figure*}

}

\section{Understanding Budget-Dependent Teacher Preference} 
\label{sec:2_obs}

In this section, we study why teacher preference differs across data budgets. Throughout, we consider vanilla KD (i.e., logit-based), where the student is trained to match the teacher's output probability \citep{hinton2015distilling}. We establish the capacity--budget crossover through controlled experiments (\Cref{sec:2_1_cap_bud}), and identify which supervision components drive the results (\Cref{sec:2_2_rank_mag}).

\subsection{Capacity--Budget Crossover in Knowledge Distillation}
    \label{sec:2_1_cap_bud}

    \begin{wrapfigure}{r}{0.43\textwidth}
    \centering

    \includegraphics[
        width=\linewidth
    ]{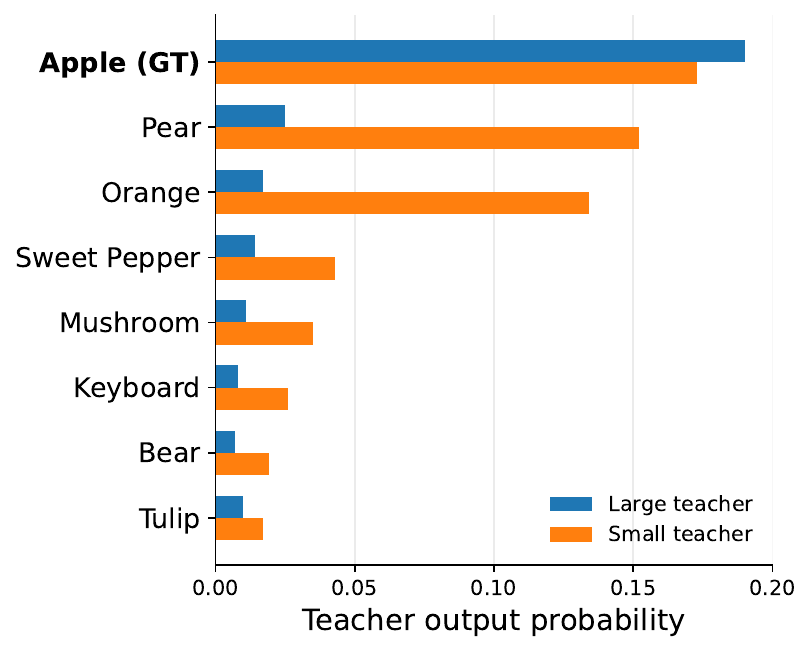}
    \caption{
        Teacher output distributions for a representative CIFAR-100 sample with temperature $T=4$.
    }
    \label{fig:teacher_output}

\end{wrapfigure}

    We first establish the capacity--budget crossover shown in \Cref{fig:overview_a}: Larger teachers become more effective as the data budget increases, whereas their advantage diminishes as the data budget decreases, resulting in a crossover at an intermediate point. Under 90\% random data pruning (i.e., randomly selecting a subset of the training data), the smaller teacher improves student accuracy by 15.56 pp, despite having substantially lower test accuracy than the larger teacher (67.44\% vs. 79.71\%; \Cref{tab:app_res_c100_teacher_cap_kd}) and approximately 63.3$\times$ fewer parameters. We observe that the same phenomenon holds across other teacher pairs in \Cref{sec:app_teacher_sweep_kd}.
    
    To understand why this capacity--budget crossover occurs, we first examine the sample-wise distribution of teacher supervision. \Cref{fig:teacher_output} illustrates a representative difference between the outputs of the large and small teachers. The large teacher spreads its non-target probability more uniformly across classes, whereas the small teacher assigns substantially more probability to semantically related classes such as `Pear' and `Orange'. This example suggests that reducing teacher capacity changes not only prediction confidence but also the structure of the supervision, providing stronger non-target signals and potentially richer dark knowledge for the student. While the contrasting output statistics in  \Cref{fig:teacher_output} provide an intuitive explanation for this behavior, they do not establish which components of teacher supervision actually cause the transferability gap.    

\subsection{Decomposing Capacity-Dependent Teacher Supervision}
    \label{sec:2_2_rank_mag}
    
\begin{wraptable}{r}{0.48\linewidth}
    \centering
    \caption{
    \textbf{Decomposition of teacher supervision into relational ordering and score geometry on CIFAR-100.} Large and small denote ResNet-34 teachers with width-64 and width-8.
    }
    \label{tab:3_rank_geometry_decomposition}

    \footnotesize
    \renewcommand{\arraystretch}{1.05}
    \setlength{\tabcolsep}{2.5pt}

    \begin{tabular}{cccc}
    \toprule
    \multirow{2}{*}{\shortstack{Relational\\Ordering}} &
    \multirow{2}{*}{\shortstack{Score\\Geometry}} &
    \multicolumn{2}{c}{Drop Rate} \\
    \cmidrule(lr){3-4}
    & & 0.1 & 0.9 \\
    \midrule

    Large & Large
    & 78.82
    & 46.93 \\

    Large & Small
    & 78.62 {\scriptsize\textcolor{red}{(-0.20)}}
    & 55.11 {\scriptsize\textcolor{green!50!black}{(+8.18)}} \\

    Small & Large
    & 78.97 {\scriptsize\textcolor{green!50!black}{(+0.15)}}
    & 50.75 {\scriptsize\textcolor{green!50!black}{(+3.82)}} \\

    Small & Small
    & 75.33 {\scriptsize\textcolor{red}{(-3.49)}}
    & 62.49 {\scriptsize\textcolor{green!50!black}{(+15.56)}} \\

    \bottomrule
    \end{tabular}

\end{wraptable}
    Motivated by these differences in teacher outputs, we quantify the contribution of each component of teacher supervision to the budget-dependent shift in teacher preference. As summarized in \Cref{fig:overview_b}, we decompose teacher outputs into two components: \textit{Relational ordering}, defined as the permutation of classes induced by sorting their outputs, and \textit{score geometry}, defined as the sorted vector of logit values. We construct hybrid logits by swapping relational ordering and score geometry between the large and small teachers, isolating the effect of each component. 
    
    \Cref{tab:3_rank_geometry_decomposition}   reports the results, with the low-data case highlighted in \Cref{fig:overview_c}. Replacing the large teacher's relational ordering with that of the small teacher improves accuracy even at a low drop rate ($+0.15$ pp), with a larger gain under aggressive pruning ($+3.82$ pp). This is consistent with \Cref{sec:2_1_cap_bud}, where the large teacher concentrates more probability on the target class, potentially weakening the relative structure among non-target classes. Score geometry, in contrast, exhibits a clear budget-dependent reversal: the large teacher performs better at a drop rate of $0.1$, whereas the small teacher improves accuracy by $8.18$ pp at $0.9$. Replacing both components yields a $15.56$ pp gain. Thus, relational ordering provides a generally useful signal from the small teacher, while score geometry is more directly tied to the capacity--budget crossover. This motivates examining how preferred sample difficulty changes with the data budget.

\section{What Should a Pruned Dataset Preserve?}{
\label{sec:3_data_sel}

    In this section, we investigate how the criteria   for selecting KD samples change as the data budget varies, focusing on two aspects: \textit{Sample-level suitability} and \textit{set-level supervision complementarity.}

    \begin{figure}[t]
    \centering

    \begin{subfigure}[t]{0.55\linewidth}
        \centering
        \includegraphics[width=\linewidth]{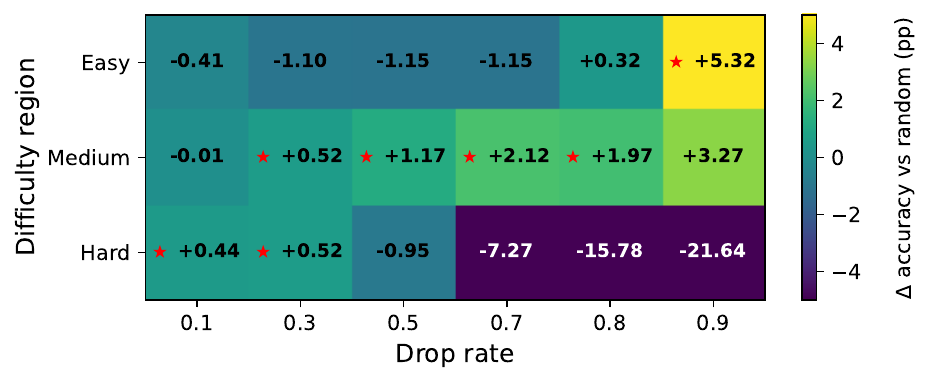}
        \caption{Sample difficulty}
        \label{fig:3_difficulty_budget}
    \end{subfigure}
    \hfill
    \begin{subfigure}[t]{0.40\linewidth}
        \centering
        \includegraphics[width=\linewidth]{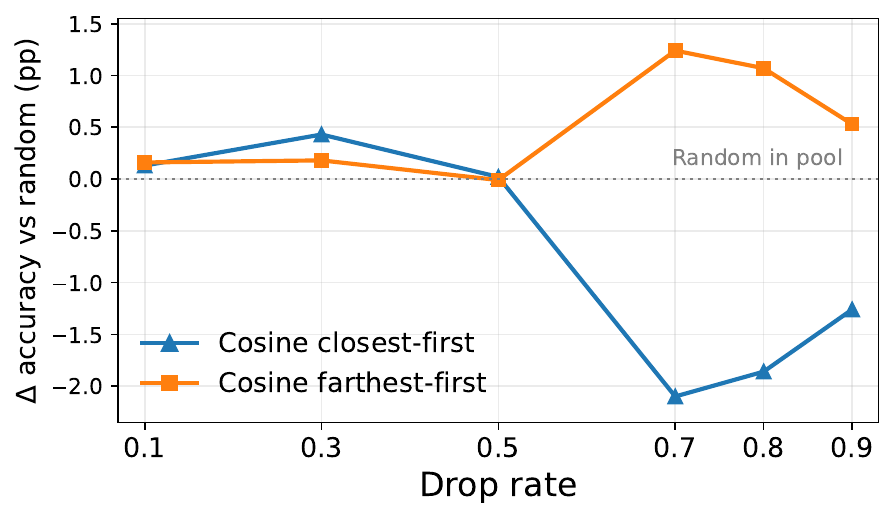}
        \caption{Supervision diversity}
        \label{fig:3_cosine_diversity}
    \end{subfigure}

    \caption{
   \textbf{Effects of sample difficulty and supervision diversity in KD.} (\textbf{a}) Samples are partitioned into easy, medium, and hard regions. Each cell reports the KD accuracy gain over random selection, with stars marking the best-performing region at each drop rate. (\textbf{b}) Comparison of selection rules within the same difficulty-filtered candidate pool, showing how diversity affects KD performance.
   }
    \label{fig:3_difficulty_diversity}
\end{figure}
    
    \textbf{Sample-level selection by budget-appropriate difficulty.} A natural question in data pruning is which samples are individually useful under a limited training budget. Here, sample difficulty is measured by the model's confidence in the ground-truth class, with lower confidence indicating harder examples; formally defined in \Cref{eq:difficulty}. Consistent with the budget-dependent difficulty effect observed in prior works \citep{sorscher2022beyond,zheng2023coverage,maharana2024d2,cho2025lightweight}, we also find that the preferred sample difficulty changes substantially with the drop rate. As shown in \Cref{fig:3_difficulty_budget}, harder examples are favored in the high-data regime, whereas the optimum gradually shifts toward medium-difficulty samples as the budget decreases. In the low-data regime, easy examples become the most effective.

    One intuitive explanation is that, in the high-data regime, representative examples are already well covered, allowing difficult samples to provide additional information for refining the decision boundary. Under the low-data regime, in contrast, allocating too much of the limited budget to difficult examples can sacrifice basic class representativeness, making easier samples relatively more useful. Thus, rather than adopting a fixed notion of sample importance, effective pruning should account for the interaction between sample difficulty and the available data budget.

    \textbf{Beyond difficulty: Set-level relational diversity.}
    Selecting samples of appropriate difficulty does not by itself guarantee an effective subset, since difficulty evaluates each sample independently. Even among individually suitable samples, many examples may provide similar teacher supervision and thus become redundant at the set level. This raises two questions: Are such redundant samples harmful to KD, and should we instead favor samples with more diverse teacher-output patterns?
    
    To probe this, we compare two controlled pairwise rules within the same difficulty-filtered candidate pool, using \emph{random} selection from that filtered pool as the reference. Given normalized teacher's output vectors $u$, let $\mathcal{A}$ denote the selected subset, and let $i$ and $j$ be the indices of the selected and candidate samples, respectively. The \emph{cosine closest-first} rule repeatedly selects the candidate most similar to the current subset, i.e.,
    \begin{equation}
         j^* = \arg\max_{j\notin \mathcal{A}} \max_{i\in 
         \mathcal{A}}\cos(u_j,u_i),
    \label{eq:cos_closest}
    \end{equation}
    whereas the \emph{cosine farthest-first} rule reverses this criterion, i.e.,
    \begin{equation}
        j^* = \arg\min_{j\notin \mathcal{A}} \max_{i\in \mathcal{A}}\cos(u_j,u_i),
    \label{eq:cos_farthest}
    \end{equation}
    where $\cos(\cdot,\cdot)$ denotes the cosine similarity between two vectors. \Cref{fig:3_cosine_diversity} reports the accuracy gain of each rule over random selection within the same difficulty-filtered pool. In the low-data regime, cosine farthest-first consistently improves KD accuracy (up to $+1.2$ points), whereas closest-first degrades it (down to $-2.1$ points). This contrast does not persist as the data budget increases: At a 50\% drop rate, both rules are within $0.02$ points of random selection, and at lower drop rates the advantage of farthest-first vanishes while closest-first can even become slightly favorable. These results suggest that, in the low-data regime, avoiding redundant supervision is important: Selecting samples with more diverse teacher-output patterns improves KD performance, while repeatedly selecting similar patterns can be harmful.
    
    Together with the difficulty analysis above, these observations suggest two complementary criteria for data selection: samples should be individually suitable for the available budget, while the selected set should avoid redundant teacher supervision. The cosine-based rules provide a simple way to probe the latter, but they rely only on greedy pairwise similarity between samples and do not directly characterize the structure of the selected set as a whole. This motivates a set-level criterion that accounts for how the selected samples collectively represent the teacher's relational space. We develop such a criterion in \Cref{sec:4_method}.

} 

\section{DVA: Difficulty- and Volume-Aware Data Selection} {
\label{sec:4_method}

Building on the observations in \Cref{sec:2_obs,sec:3_data_sel}, we propose DVA (Difficulty- and Volume-Aware data selection). Although smaller teachers can be preferable under aggressive pruning, replacing the distillation teacher is not always practical: the teacher is often fixed (e.g., off-the-shelf checkpoints), and the crossover point varies across settings (\Cref{sec:app_teacher_sweep_kd}). We therefore use the small model not as the distillation teacher, but as a selection proxy while retaining the given teacher for final distillation.

DVA consists of three stages. First, we construct a candidate pool of data with samples having difficulty appropriate for the given data budget. Second, we characterize the class-conditional structure of teacher outputs through their principal relational directions and corresponding importance. Finally, we select a subset that maximizes set-level relational diversity by favoring examples whose teacher outputs span a large volume in the class-conditional proxy-output space.

\textbf{Formalism: Data pruning for KD.} Let $\mathcal{S}=\{(x_i,y_i)\}_{i=1}^{N}$ denote the training set with $N$ samples, where $x_i$ is the $i$-th training sample and $y_i$ is its class label. Let $\mathcal{S}_c=\{(x_i,y_i)\in\mathcal{S}\mid y_i=c\}$ denote the samples belonging to class $c\in\{1,\ldots,C\}$. Furthermore, let $\rho\in[0,1)$ denote the drop rate, defined as the fraction of training samples to be removed from the original training set.

Let $T_L$ be the (large) teacher that we use for student distillation. Then, the goal of data pruning can be stated as finding a subset $\mathcal{M} \subseteq \mathcal{S}$ that maximizes the downstream KD performance, i.e.,
\begin{equation}
    \mathcal{M}^*(\rho) = \arg\max_{\substack{ \mathcal{M}\subseteq\mathcal{S}\\ |\mathcal{M}|=\lfloor(1-\rho)N\rfloor }} \mathsf{KDPerf}(\mathcal{M};T_L),
\end{equation}
where $\mathrm{KDPerf}(\mathcal{M};T_L)$ denotes the student accuracy obtained by distilling from $T_L$ on $\mathcal{M}$. Directly optimizing this objective is impractical because evaluating $\mathsf{KDPerf}$ requires training the student model for each candidate subset $\mathcal{M}$. The idea of DVA is to ease this search, by utilizing the predictions of a small teacher $T_s$ as proxy signals for data selection.

\textbf{Stage 1: Budget-aware difficulty filtering.} The first step of DVA is to construct a class-wise candidate pool $\mathcal{P}_c \subseteq \mathcal{S}_c$ which contains only the samples with appropriate difficulty, corresponding to the training budget. More concretely, we define the difficulty of sample $x_i$ using the probability assigned by the small teacher $T_s$ to the ground-truth class $y_i$, as
\begin{align}
    d_i = 1 - p_{T_s}(y_i \mid x_i).
\label{eq:difficulty}
\end{align}
Here, a larger $d_i$ indicates a more difficult sample. We also define the quantity $p_i := p_{T_s}(\cdot~|~x_i) \in \mathbb{R}^C$, which is the full class probability vector predicted by the proxy model $T_s$.

Given this difficulty, we construct $\mathcal{P}_c$ by retaining the samples having the difficulty within an interval $d_i \in \mathcal{I}$, where this interval $\mathcal{I}$ is determined based on the data budget. We set the interval size to $((1-\rho)+0.1)|\mathcal{S}_c|$ for $\rho\in\{0.3,0.5\}$, and to $2(1-\rho)|\mathcal{S}_c|$ for $\rho\in\{0.7,0.8,0.9\}$. We describe the detailed procedure in \Cref{app:diff_filt}.

We note that we constrain $\mathcal{P}_c$ to satisfy $|\mathcal{P}_c| \geq k_c := \lfloor(1-\rho)|\mathcal{S}_c|\rfloor$. This ensures that we retain sufficient samples from each class to achieve a uniform drop rate over the classes, so that the data pruning may not lead to a highly class-imbalanced training.

\textbf{Stage 2: Relational projection.}
Given the candidate pool $\mathcal{P}_c$, we construct a representation for set-level diversity. Unlike cosine similarity (\Cref{sec:3_data_sel}), which captures pairwise direction but ignores variation across directions, we characterize the full class-conditional proxy-output distribution by its principal directions and relative scales, computed over $\mathcal{S}_c$.

Precisely, for each class $c$, we compute the mean proxy output as $\mu_c = \frac{1}{|\mathcal{S}_c|} \sum_{(x_i,y_i)\in\mathcal S_c} p_i$, and stack the centered proxy outputs as rows of $X_c \in \mathbb{R}^{|\mathcal{S}_c| \times C}$, whose $i$-th row is $(p_i - \mu_c)^\top$ for $(x_i, y_i) \in \mathcal{S}_c$. We perform singular value decomposition (SVD) on $X_c$,
\begin{equation}
    X_c = U_c \Sigma_c V_c^\top,
\label{eq:class_svd}
\end{equation}
where $V_c \in \mathbb{R}^{C \times R}$, $R= \operatorname{rank}(X_c)$, contains the principal directions of the class-conditional proxy-output distribution. Here, $\Sigma_c \in \mathbb{R}^{R \times R}$ encodes the scale of variation along these directions. We then use these statistics to map each candidate sample $(x_i,y_i) \in \mathcal{P}_c$ into the weighted principal space as
\begin{equation}
    a_i =
    \frac{\Sigma_c}{\sqrt{|\mathcal{S}_c|-1}}
    V_c^\top (p_i-\mu_c).
\label{eq:weighted_projection}
\end{equation}
The representation $a_i$ captures how sample $i$ deviates from the class mean along each direction, with coordinates weighted by the amount of class-conditional variation captured by that direction. \Cref{sec:app_principal_directions} shows that the dominant directions correspond to recurring confusions with competing classes.

\textbf{Stage 3: Volume maximization.}
After projecting each candidate into the weighted relational space, we seek a subset that covers important relational directions while avoiding redundant proxy outputs. Measuring only the subspace spanned by the selected samples is insufficient, since it ignores the magnitude of variation along each direction and becomes uninformative once the selected samples span the full relational subspace. We therefore measure the volume spanned by the selected weighted representations using a log-determinant objective: 
\begin{equation}
    \mathcal{M}_c^{*} = \arg\max_{\substack{ \mathcal{M}_c\subseteq\mathcal{P}_c\\ |\mathcal{M}_c|=k_c}} \log\det \left( \delta I_R + \sum_{(x_i,y_i)\in\mathcal{M}_c} a_i a_i^\top \right),
\label{eq:logdet_objective}
\end{equation}
where we set the regularization constant to $\delta=10^{-3}$. This value is small relative to the dominant variation in the projected space and primarily stabilizes the log-determinant along directions that are not yet well covered by the selected subset. Because each coordinate of $a_i$ is scaled by $\Sigma_c$, the objective emphasizes variation along important principal directions while assigning diminishing gain to directions already well covered by the selected subset.

To avoid recomputing the determinant from scratch for every candidate, we use the matrix determinant lemma \citep{harville1997matrix}. We define 
\begin{equation}
   G(\mathcal{M}_c) = \delta I_R + \sum_{(x_j,y_j)\in \mathcal{M}_c} a_j a_j^\top, 
\end{equation}
where $I_R$ denotes $R\times R$ identity matrix. Note that, with any scalar $\delta>0$, $G(\mathcal{M}_c)$ is positive definite and therefore invertible. Applying the matrix determinant lemma to the rank-one update
$G(\mathcal{M}_c) + a_i a_i^\top$ gives the marginal gain of adding candidate $i$ as
\begin{equation}
    \Delta_i = \log\det G(\mathcal{M}_c \cup \{(x_i,y_i)\}) - \log\det G(\mathcal{M}_c) = \log\!\left( 1 + a_i^\top G(\mathcal{M}_c)^{-1} a_i \right).
\end{equation}
We greedily add the candidate with the largest $\Delta_i$ until $|\mathcal{M}_c|=k_c$. After each selection, we update $G(\mathcal{M}_c)^{-1}$ using the Sherman--Morrison formula~\citep{sherman1950adjustment}, avoiding repeated matrix inversion and reducing the computational cost of greedy selection.  The detailed algorithm is provided in \Cref{app:algo}.

The final selected subset is $\mathcal{M}=\bigcup_{c=1}^{C}\mathcal{M}_c$.

}

\section{Experiment} {
\label{sec:5_exp}
\begin{table}[t]

\centering
\caption{
\textbf{KD accuracy (\%) on CIFAR-100 with a ResNet-18 student.}
All baselines use the ResNet-34 teacher with width-64 for KD unless otherwise specified.
Results are averaged over 10 seeds and reported as mean $\pm$ standard deviation. Best/second-best results are in \textbf{bold}/\underline{underline}.
}
\label{tab:cifar100_main}

\footnotesize
\renewcommand{\arraystretch}{1.10}
\setlength{\tabcolsep}{5pt}

\resizebox{0.85\linewidth}{!}{%
\begin{tabular}{l cccccc}
\toprule
\multirow{2}{*}{Method} &
\multicolumn{6}{c}{Drop Rate} \\
\cmidrule(lr){2-7}
& 0.1 & 0.3 & 0.5 & 0.7 & 0.8 & 0.9 \\
\midrule
\multicolumn{7}{l}{\textit{Training-Dynamics-Based}} \vspace{1pt} \\

Forgetting
& \textbf{79.35}{\scriptsize$\pm$0.27}
& 77.69{\scriptsize$\pm$0.29}
& 73.56{\scriptsize$\pm$0.37}
& 61.45{\scriptsize$\pm$0.22}
& 51.02{\scriptsize$\pm$0.51}
& 35.06{\scriptsize$\pm$0.73} \\

EL2N
& 79.06{\scriptsize$\pm$0.28}
& 77.21{\scriptsize$\pm$0.25}
& 70.23{\scriptsize$\pm$0.30}
& 50.85{\scriptsize$\pm$0.68}
& 35.02{\scriptsize$\pm$0.60}
& 16.44{\scriptsize$\pm$0.29} \\

CCS
& 79.13{\scriptsize$\pm$0.23}
& {\ul 77.81}{\scriptsize$\pm$0.23}
& 74.81{\scriptsize$\pm$0.21}
& 69.82{\scriptsize$\pm$0.36}
& 64.84{\scriptsize$\pm$0.35}
& 53.77{\scriptsize$\pm$0.62} \\

D$^2$
& 79.02{\scriptsize$\pm$0.27}
& 77.58{\scriptsize$\pm$0.20}
& {\ul 74.96}{\scriptsize$\pm$0.17}
& {\ul 69.99}{\scriptsize$\pm$0.27}
& \textbf{66.14}{\scriptsize$\pm$0.34}
& {\ul 54.49}{\scriptsize$\pm$0.63} \\

DUAL
& {\ul 79.29}{\scriptsize$\pm$0.28}
& \textbf{77.99}{\scriptsize$\pm$0.15}
& \textbf{75.44}{\scriptsize$\pm$0.18}
& \textbf{70.47}{\scriptsize$\pm$0.13}
& {\ul 65.86}{\scriptsize$\pm$0.24}
& \textbf{55.56}{\scriptsize$\pm$0.45} \\

\midrule

\multicolumn{7}{l}{\textit{Training-Dynamics-Free}} \vspace{1pt} \\

Random
& 78.82{\scriptsize$\pm$0.25}
& 76.69{\scriptsize$\pm$0.17}
& {\ul 73.16}{\scriptsize$\pm$0.20}
& 67.07{\scriptsize$\pm$0.49}
& 61.35{\scriptsize$\pm$0.43}
& 46.93{\scriptsize$\pm$0.80} \\

GraNd 
& 78.64{\scriptsize$\pm$0.32}
& 75.28{\scriptsize$\pm$0.23}
& 69.66{\scriptsize$\pm$0.18}
& 59.01{\scriptsize$\pm$0.29}
& 48.97{\scriptsize$\pm$0.38}
& 30.12{\scriptsize$\pm$0.44} \\

MDSLR
& 78.58{\scriptsize$\pm$0.25}
& 76.34{\scriptsize$\pm$0.21}
& 73.11{\scriptsize$\pm$0.24}
& {\ul 67.17}{\scriptsize$\pm$0.26}
& 62.25{\scriptsize$\pm$0.27}
& 48.10{\scriptsize$\pm$0.76} \\

IF-Beta
& {\ul 79.05}{\scriptsize$\pm$0.21}
& {\ul 76.93}{\scriptsize$\pm$0.19}
& 71.21{\scriptsize$\pm$0.27}
& 64.62{\scriptsize$\pm$0.46}
& {\ul 63.25}{\scriptsize$\pm$0.27}
& {\ul 48.99}{\scriptsize$\pm$0.75} \\

\rowcolor{gray!12} DVA (Ours)
& \textbf{79.13}{\scriptsize$\pm$0.17}
& \textbf{77.43}{\scriptsize$\pm$0.20}
& \textbf{74.62}{\scriptsize$\pm$0.16}
& \textbf{69.09}{\scriptsize$\pm$0.23}
& \textbf{64.45}{\scriptsize$\pm$0.36}
& \textbf{53.76}{\scriptsize$\pm$0.77} \\

\bottomrule
\end{tabular}
}

\end{table}

We evaluate our method from four perspectives. First, we compare it with existing data pruning methods under standard single-teacher KD (\Cref{sec:5_main}). Second, we disentangle the contribution of each stage of our framework (\Cref{sec:5_effect}). Third, we examine the effect of using a small teacher as the selection proxy (\Cref{sec:5_proxy}). Finally, we broaden our evaluation across datasets and model architectures (\Cref{sec:5_broader}), including Tiny-ImageNet, ViT-based KD, and ImageNet-scale experiments. We further examine our findings under a broader range of settings, including multi-teacher supervision and additional cross-teacher and robustness analyses, in \Cref{sec:app_broad_exp,sec:app_mtkd,sec:app_cross_kd}.

\subsection{Experimental Setup}
\label{sec:5_setup}

\textbf{Datasets and models.} We primarily conduct controlled experiments on CIFAR-100. Unless otherwise specified, we use a ResNet-34 teacher and a ResNet-18 student for KD. To study the effect of teacher capacity, we construct a family of ResNet-34 teachers by varying the network width while keeping the overall architecture fixed. This setup allows us to isolate the interaction between teacher capacity and the available student-training budget. We further evaluate the generality of our observations across architectures using ViTs on CIFAR-100, and across dataset scales on Tiny-ImageNet \citep{le2015tiny} and ImageNet \citep{deng2009imagenet}.

\textbf{Baselines.} We primarily compare our method with training-dynamics-free data pruning methods, including GraNd~\citep{paul2021deep}, MDSLR~\citep{chen2025medium} and IF-Beta~\citep{wu2026distill}, as these methods share our goal of constructing subsets without relying on student training dynamics. For broader comparison, we also report training-dynamics-based methods, including EL2N~\citep{paul2021deep}, Forgetting~\citep{toneva2019empirical}, CCS~\citep{zheng2023coverage}, D$^2$~\citep{maharana2024d2}, and DUAL~\citep{cho2025lightweight}, which leverage statistics collected during model training.

Detailed baseline implementations and additional experimental settings are provided in \Cref{sec:app_det_exp_set}.

\subsection{Main Experiment}
\label{sec:5_main}

\Cref{tab:cifar100_main} compares our method with training-dynamics-based and training-dynamics-free data pruning methods on CIFAR-100. From the table, our method achieves the best accuracy among training-dynamics-free approaches across all drop rates, with the advantage becoming more pronounced as the data budget decreases. At a $0.9$ drop rate, it reaches $53.76\%$, outperforming the strongest training-dynamics-free baseline, IF-Beta, by $4.77$ pp. Despite not relying on student training dynamics for subset construction, our method also remains competitive with strong training-dynamics-based approaches such as CCS, D$^2$, and DUAL. These results indicate that explicitly modeling the relational structure of proxy outputs substantially improves data selection for KD, especially in the low-data regimes (e.g., $0.7$, $0.8$, and $0.9$ drop rates). 

\subsection{Effect of Difficulty Filtering and Volume Maximization}
\label{sec:5_effect}

For more detailed analysis of our method, we first disentangle the contributions of difficulty filtering and of volume maximization. To isolate the effect of difficulty filtering, we construct \textit{filtered random}, which randomly selects the final subset from exactly the same difficulty-filtered candidate pool used by our method. Comparing random with filtered random therefore measures the benefit of restricting selection to a budget-appropriate difficulty region, while comparing filtered random with Ours isolates the additional contribution of volume maximization.

\begin{wraptable}{r}{0.55\linewidth}
\centering
\caption{
\textbf{Effect of difficulty filtering and volume maximization.}
Gain is measured as the difference between Ours and Filtered Random, averaged over 10 seeds.
}
\label{tab:filtered_random}

\footnotesize
\renewcommand{\arraystretch}{1.05}
\setlength{\tabcolsep}{2.8pt}

\begin{tabular}{ccccc}
\toprule
Drop & Random & Filtered Rand. & DVA (Ours) & Gain \\
\midrule

0.7
& 67.07{\scriptsize\, $\pm$ 0.49}
& 68.09{\scriptsize\, $\pm$ 0.20}
& \textbf{69.09}{\scriptsize\, $\pm$ 0.23}
& \textbf{+1.00} \\

0.8
& 61.35{\scriptsize\, $\pm$ 0.43}
& 63.40{\scriptsize\, $\pm$ 0.32}
& \textbf{64.45}{\scriptsize\, $\pm$ 0.36}
& \textbf{+1.05} \\

0.9
& 46.93{\scriptsize\, $\pm$ 0.80}
& 53.02{\scriptsize\, $\pm$ 0.62}
& \textbf{53.76}{\scriptsize\, $\pm$ 0.77}
& \textbf{+0.74} \\

\bottomrule
\end{tabular}

\end{wraptable}

As shown in \Cref{tab:filtered_random}, difficulty filtering alone provides increasing gains as the data budget becomes more constrained, improving KD accuracy by $1.02$, $2.05$, and $6.09$ points at drop rates of $0.7$, $0.8$, and $0.9$, respectively. On top of this improvement, our relational selection further increases accuracy by $1.00$, $1.05$, and $0.74$ points at the corresponding drop rates. These results indicate that budget-aware difficulty filtering accounts for a substantial portion of the improvement in the low-data regime, while set-level relational selection also provides a consistent gain within the filtered candidate pool. We further verify the contribution of the volume-maximization step in \Cref{sec:app_volume_search} by comparing it with low-volume selection while keeping the filtered candidate pool fixed.

\subsection{Effect of Small Teacher as Proxy}
\label{sec:5_proxy}

\begin{table}[t]
\centering
\caption{
\textbf{Effect of proxy teacher capacity in DVA on CIFAR-100.} We vary the width of the ResNet-34 proxy teacher from 4 to 64 while fixing the distillation teacher to the ResNet-34 with width-64. All runs are averaged over 10 seeds. Best/second-best results are in \textbf{bold}/\underline{underline}.}
\label{tab:proxy_teacher_capacity}
\resizebox{0.95\linewidth}{!}{
\begin{tabular}{lcccccc}
\toprule
& \multicolumn{6}{c}{Drop Rate} \\
\cmidrule(lr){2-7}
Proxy Teacher
& 0.1 & 0.3 & 0.5 & 0.7 & 0.8 & 0.9 \\
\midrule

Width-4
& \textbf{79.14}{\scriptsize$\pm$0.19}
& \underline{77.06}{\scriptsize$\pm$0.28}
& \underline{73.84}{\scriptsize$\pm$0.26}
& \underline{67.85}{\scriptsize$\pm$0.18}
& \underline{63.17}{\scriptsize$\pm$0.30}
& \underline{53.31}{\scriptsize$\pm$0.67} \\

Width-8 (small)
& \underline{79.13}{\scriptsize$\pm$0.17}
& \textbf{77.43}{\scriptsize$\pm$0.20}
& \textbf{74.62}{\scriptsize$\pm$0.16}
& \textbf{69.09}{\scriptsize$\pm$0.23}
& \textbf{64.45}{\scriptsize$\pm$0.36}
& \textbf{53.76}{\scriptsize$\pm$0.77} \\

Width-16
& 79.09{\scriptsize$\pm$0.22}
& 76.99{\scriptsize$\pm$0.18}
& 73.77{\scriptsize$\pm$0.27}
& 67.91{\scriptsize$\pm$0.20}
& 63.05{\scriptsize$\pm$0.32}
& 50.30{\scriptsize$\pm$1.19} \\

Width-32
& 79.00{\scriptsize$\pm$0.22}
& 76.74{\scriptsize$\pm$0.33}
& 73.42{\scriptsize$\pm$0.26}
& 65.94{\scriptsize$\pm$0.29}
& 59.48{\scriptsize$\pm$0.72}
& 42.05{\scriptsize$\pm$1.38} \\

Width-64 (large)
& 79.12{\scriptsize$\pm$0.28}
& 76.62{\scriptsize$\pm$0.20}
& 73.25{\scriptsize$\pm$0.45}
& 64.09{\scriptsize$\pm$0.25}
& 55.45{\scriptsize$\pm$0.63}
& 35.96{\scriptsize$\pm$1.14} \\

\midrule
$\Delta$ (Width-8 $-$ Width-64)
& +0.01
& +0.81
& +1.37
& +5.00
& +9.00
& +17.80 \\

\bottomrule
\end{tabular}
}
\end{table}

We next examine how the capacity of the selection proxy affects DVA by varying the width of the ResNet-34 proxy teacher from 4 to 64 while keeping the selection procedure fixed. As shown in \Cref{tab:proxy_teacher_capacity}, proxy capacity has little effect at low pruning rates, but lower-capacity proxies become increasingly favorable as pruning becomes more aggressive. At a drop rate of $0.9$, accuracy decreases from $53.76\%$ with the width-8 proxy to $50.30\%$, $42.05\%$, and $35.96\%$ with width-16, width-32, and width-64 proxies, respectively. In contrast, at a drop rate of $0.1$, all proxy widths achieve comparable performance around $79\%$. This widening gap under aggressive pruning indicates that the choice of selection proxy becomes particularly important in the low-data regime, with lower-capacity teachers providing more effective signals for DVA.

\subsection{Broader Experiments}
\label{sec:5_broader}

\begin{figure*}[t]

    \centering

    \begin{subfigure}[t]{0.44\textwidth}
        \centering
        \includegraphics[
            width=\linewidth,
        ]{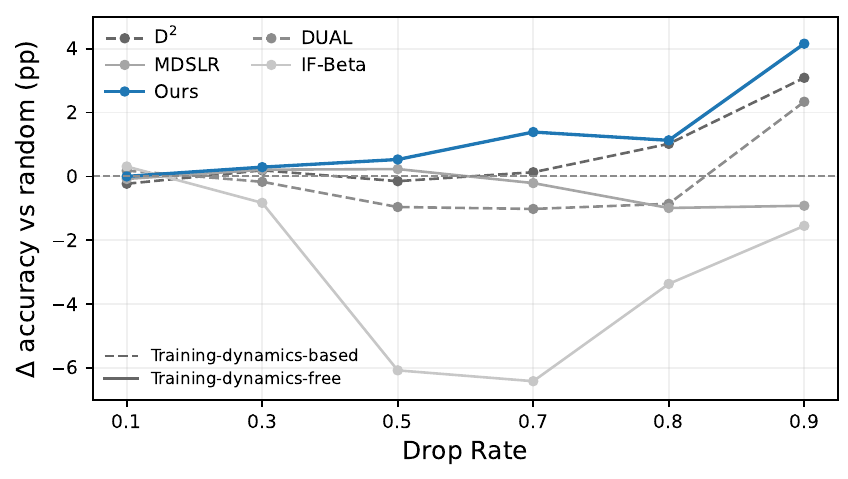}
        \caption{CIFAR-100, ViT, Single Teacher}
        \label{fig:gain_vit_single}
    \end{subfigure}
    \hfill
    \begin{subfigure}[t]{0.44\textwidth}
        \centering
        \includegraphics[
            width=\linewidth,
        ]{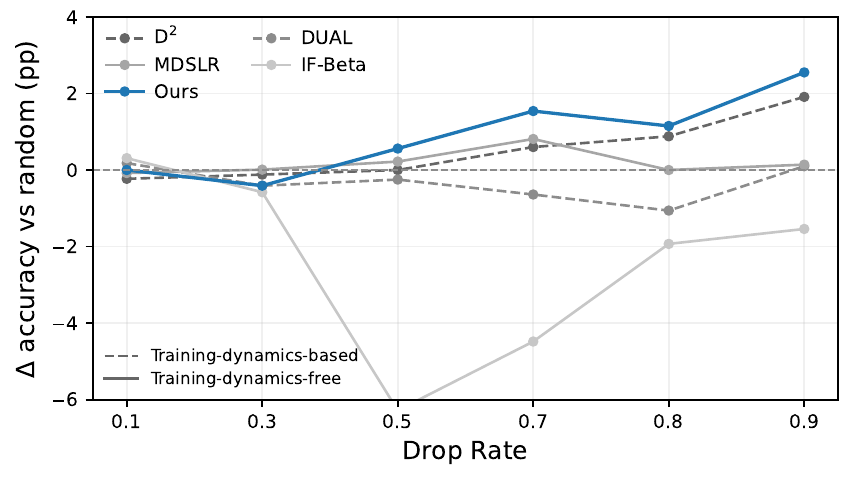}
        \caption{CIFAR-100, ViT, Multi-Teacher}
        \label{fig:gain_vit_multi}
    \end{subfigure}

    \begin{subfigure}[t]{0.44\textwidth}
        \centering
        \includegraphics[
            width=\linewidth,
        ]{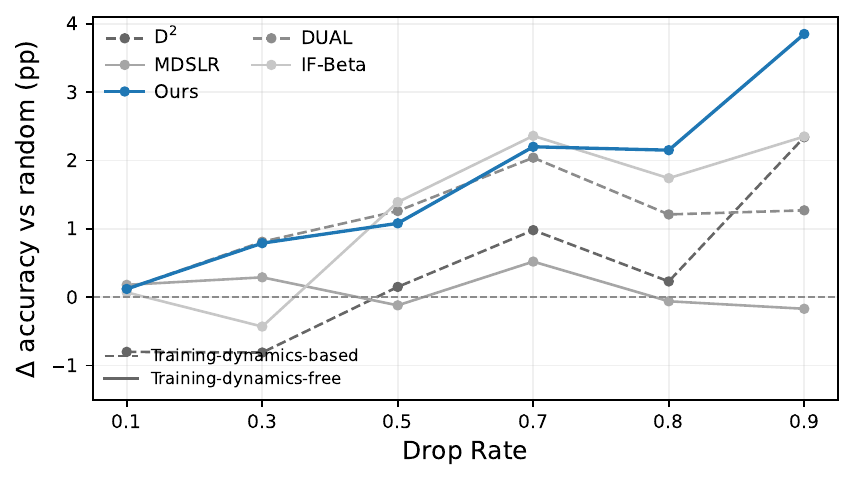}
        \caption{Tiny-ImageNet, ResNet, Single Teacher}
        \label{fig:gain_tiny_single}
    \end{subfigure}
    \hfill
    \begin{subfigure}[t]{0.44\textwidth}
        \centering
        \includegraphics[
            width=\linewidth,
        ]{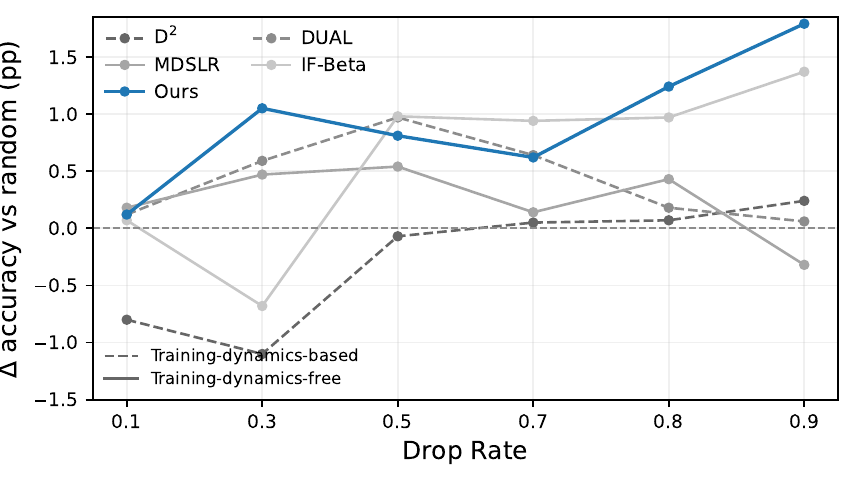}
        \caption{Tiny-ImageNet, ResNet, Multi-Teacher}
        \label{fig:gain_tiny_multi}
    \end{subfigure}

    \caption{
    \textbf{Accuracy gain over random selection across additional experimental settings.}
    Each panel reports the KD accuracy difference relative to random selection at each drop rate, averaged over 3 seeds. Our method is highlighted in blue, while competing methods are shown in grayscale.
    }
    \label{fig:5_vit_tin}

\end{figure*}


\textbf{Generalization across datasets and architectures.}
We first evaluate our method beyond the default ResNet on CIFAR-100 setting. \Cref{fig:5_vit_tin} reports the accuracy gain over random selection at ViT on CIFAR-100, and ResNet on Tiny-ImageNet, under both single-teacher and scheduled multi-teacher supervision. Across these settings, our method generally provides larger gains in the low-data regime, even outperforming strong training-dynamics-based methods. Notably, our method also performs strongly under multi-teacher supervision. These results show that the proposed selection principle is not specific to a particular architecture, dataset, or single-teacher setting.

\textbf{Scaling to ImageNet.}
We further evaluate our method at ImageNet scale. Unlike the controlled CIFAR-100 experiments, where teacher variants are explicitly trained to study model capacity, we directly use publicly available pretrained teacher checkpoints \textit{without retraining or adapting} them for our selection method. For this experiment, we consider teacher models that vary in network depth rather than width (ResNet-10, ResNet-34), allowing us to examine whether our findings extend to a different form of model capacity scaling. We use ResNet-10 as the selection proxy and ResNet-34 as the teacher in the subsequent distillation stage.

\begin{wraptable}{r}{0.50\linewidth}
\centering
\caption{
\textbf{KD accuracy (\%) on ImageNet with a ResNet-18 student distilled from
a ResNet-34 teacher.}
Results are averaged over 3 seeds. Best/second-best results are in \textbf{bold}/\underline{underline}.
}
\label{tab:imagenet_resnet_main}
\vspace{-8pt}
\footnotesize
\renewcommand{\arraystretch}{0.95}
\setlength{\tabcolsep}{3.2pt}
\resizebox{\linewidth}{!}{%
\begin{tabular}{lcccccc}
\toprule
\multirow{2}{*}{Method} &
\multicolumn{6}{c}{Drop Rate} \\
\cmidrule(lr){2-7}
& 0.1 & 0.3 & 0.5 & 0.7 & 0.8 & 0.9 \\
\midrule

\multicolumn{7}{l}{\textit{Training-Dynamics-Based}} \\

D$^2$
& {\ul 71.42}
& {\ul 69.90}
& {\ul 67.68}
& {\ul 65.01}
& \textbf{61.58}
& \textbf{52.79} \\

DUAL
& \textbf{71.67}
& \textbf{71.15}
& \textbf{69.40}
& \textbf{65.53}
& {\ul 61.40}
& {\ul 52.18} \\

\midrule

\multicolumn{7}{l}{\textit{Training-Dynamics-Free}} \\

Random
& 71.32
& 70.24
& 68.49
& 64.31
& 59.96
& 50.75 \\

MDSLR
& 71.49
& {\ul 70.53}
& {\ul 68.76}
& 64.42
& 60.11
& 51.01 \\

IF-Beta
& \textbf{71.64}
& 70.46
& 68.75
& {\ul 65.17}
& {\ul 61.09}
& {\ul 51.73} \\

\rowcolor{gray!12} DVA (Ours)
& {\ul 71.60}
& \textbf{70.97}
& \textbf{68.81}
& \textbf{65.30}
& \textbf{61.23}
& \textbf{52.47} \\

\bottomrule
\end{tabular}
}
\end{wraptable}

As shown in \Cref{tab:imagenet_resnet_main}, our method consistently improves over random selection and achieves the best accuracy among training-dynamics-free methods at five of the six pruning rates. The improvement becomes more pronounced in the low-data regime, increasing from $0.28$ pp at a $0.1$ drop rate to $1.72$ pp at a $0.9$ drop rate. In this experiment, we consider models scaled by network depth rather than width, providing an additional setting to examine whether our findings generalize across different forms of model capacity scaling. Together, these results show that our selection strategy extends to large-scale data and can be applied directly to \textit{off-the-shelf pretrained teachers}.

}

\section{Conclusion} {

In this work, we investigate KD under data pruning, with a particular focus on selecting informative training subsets without relying on student training dynamics. Our analyses show that teacher--student capacity alone does not explain KD performance under varying data budgets. That is, smaller teachers can provide more effective supervision in the low-data regime, and we reveal that both \textit{relational ordering} and \textit{score geometry} contribute to this behavior. We further identify two properties of effective pruned subsets---budget-appropriate difficulty and relational diversity. Motivated by these observations, we proposed DVA, a training-dynamics-free data selection framework built on a small-teacher proxy, that consistently improves KD performance across different pruning rates and experimental settings. Our results further show that the approach can be applied to pretrained teachers without additional teacher training or adaptation.

}

\subsection*{AI use statement}

In this work, we used generative AI tools to refine research hypotheses, provide feedback on experimental design, implement methods, and interpret experimental results. We did not use generative AI tools to generate synthetic datasets, develop theoretical models or conceptual frameworks, formulate mathematical claims, or assist in proving mathematical claims; translation, dataset cleaning and reformatting, and qualitative or thematic data analysis were not applicable to this work. Additionally, we used generative AI tools to identify and summarize related literature, create and edit research code, edit parts of the manuscript, improve readability, and format tables, figures, and references. We reviewed all AI-assisted work: all hypotheses, experimental designs, results, and claims were independently evaluated by the authors, and all AI-assisted code was verified and tested for correctness by the authors. We take responsibility for the final content of this work, including any text, claims, or artifacts produced with the aid of generative AI.

\bibliography{main}
\bibliographystyle{iclr2027_conference}

\newpage

\appendix
\appendix
\etocdepthtag.toc{appendix}

\etocsettagdepth{main}{none}
\etocsettagdepth{appendix}{subsection}
\etocobeydepthtags
\hypersetup{linkcolor=black}
\renewcommand{\contentsname}{Appendix}
\tableofcontents
\newpage
\hypersetup{linkcolor=magenta}

\section{Limitation and Future Work} {

Our framework assumes access to a suitable low-capacity proxy teacher and does not automatically determine the most appropriate teacher capacity for a given data budget. In addition, the budget-aware difficulty filtering currently relies on a pre-defined mapping between pruning rate and preferred difficulty region. Automatically selecting the proxy teacher and adapting the difficulty regime from the teacher-output distribution remain open directions.

Future work should examine whether the same capacity--budget relationship and relational selection principles extend to larger-scale distillation settings, such as language models and foundation vision encoders. Such settings may provide opportunities to reduce the data and computation required to distill large pre-trained models while preserving informative teacher supervision.

}
\section{Related Work} {

\paragraph{Teacher capacity in KD.}

Knowledge distillation transfers supervision from a teacher to a smaller student through the teacher's predictive distribution~\citep{hinton2015distilling}. Prior work has shown that a more accurate or larger teacher is not always more transferable: large teacher--student capacity gaps can hinder distillation~\citep{cho2019efficacy}, motivating intermediate teacher assistants~\citep{mirzadeh2020improved}. More recently, \citet{benbaruch2025distilling} observed that teacher capacity also interacts with data pruning, with smaller teachers outperforming larger ones in the low-data regime.

Other studies have examined how target and non-target supervision should be transferred. \citet{zhao2022decoupled} decouple target- and non-target-class knowledge, while \citet{li2022asymmetric} reshape non-target probabilities to improve their discriminability. Unlike these works, which modify how teacher supervision is transferred, we use relational ordering and score geometry as an analytical decomposition to characterize how teacher supervision interacts with the data budget. This analysis explains the capacity--budget relationship and subsequently motivates our data-selection framework.

\paragraph{Data pruning and KD-specific sample selection.}

Data pruning selects a compact subset of training samples while preserving downstream performance. Existing approaches exploit sample forgetting \citep{toneva2019empirical}, gradient- or error-based importance such as GraNd and EL2N \citep{paul2021deep}, and coverage- or training-dynamics-based criteria such as CCS, DUAL, and D$^2$ \citep{zheng2023coverage,cho2025lightweight,maharana2024d2}.

To reduce selection overhead, \citet{coleman2019selection} showed that a smaller proxy model can perform data selection for a larger target model, with selection signals transferring across model capacities. In contrast, our low-capacity proxy is motivated not merely by computational efficiency, but by our finding that lower-capacity teachers provide more effective supervision under limited data, making their outputs particularly suitable for subset selection.

Recent work has also considered data selection specifically for KD. MDSLR \citep{chen2025medium} focuses on samples near a smoothed decision boundary, while IF-Beta \citep{wu2026distill} learns a KD-oriented sampling policy from sample utility. Unlike these methods, our approach explicitly uses class-conditional proxy-output relations for subset construction, combining budget-dependent difficulty filtering with volume maximization in the proxy relational space, without relying on student training dynamics.

\paragraph{Coreset selection and diversity.} 
Coverage-aware pruning methods such as CCS \citep{zheng2023coverage} highlight the importance of avoiding redundant or poorly representative subsets under aggressive pruning. More broadly, determinant-based subset selection has long been used to favor diverse and non-redundant subsets. Determinantal point processes assign higher probability to subsets spanning larger volumes, and have also been studied directly for coreset construction  \citep{kulesza2012determinantal,tremblay2019determinantal}. Our method builds on this general principle, but performs volume maximization in a class-conditional relational space derived from teacher outputs, where principal directions are weighted according to their importance for the teacher-output distribution.

}
\section{Positioning of DVA}

\Cref{tab:app_method_comparison} summarizes the methodological differences between DVA and representative data pruning approaches. Unlike methods that rely on student training dynamics or additional optimization to obtain selection statistics, DVA can perform data selection before student training once the teacher and selection proxy are available. DVA further combines two complementary aspects of fine-grained subset construction: set-level selection, which accounts for interactions among selected samples, and budget-aware selection, which adapts the selection criterion to the available data budget. Finally, DVA is designed specifically for KD and explicitly models the class-relational geometry encoded in teacher outputs. Together, these properties distinguish DVA from prior approaches by combining training-dynamics-free selection, budget-dependent subset construction, and teacher-output relational geometry within a KD-specific framework.


\begin{table}[H]
\centering
\caption{Comparison of representative data-pruning methods.}
\label{tab:app_method_comparison}

\resizebox{\textwidth}{!}{%
\begin{tabular}{lcccccc}
\toprule
\multirow{2}{*}{Method}
& \multicolumn{2}{c}{Selection Requirement}
& \multicolumn{2}{c}{Selection Strategy}
& \multicolumn{2}{c}{KD Awareness} \\
\cmidrule(lr){2-3}
\cmidrule(lr){4-5}
\cmidrule(lr){6-7}
&
\makecell{Dynamics-\\free}
&
\makecell{No extra\\optimization}
&
\makecell{Set-level\\selection}
&
\makecell{Budget-\\aware}
&
\makecell{KD-\\specific}
&
\makecell{Teacher relational\\geometry}
\\
\midrule

Forgetting \citep{toneva2019empirical}
& \xmark & \xmark
& \xmark & \xmark
& \xmark & \xmark \\

EL2N \citep{paul2021deep}
& \xmark & \xmark
& \xmark & \xmark
& \xmark & \xmark \\

GraNd \citep{paul2021deep}
& \xmark & \xmark
& \xmark & \xmark
& \xmark & \xmark \\

CCS \citep{zheng2023coverage}
& \xmark & \xmark
& \cmark & \cmark
& \xmark & \xmark \\

D$^2$ \citep{maharana2024d2}
& \xmark & \xmark
& \cmark & \cmark
& \xmark & \xmark \\

DUAL \citep{cho2025lightweight}
& \xmark & \xmark
& \xmark & \cmark
& \xmark & \xmark \\

MDSLR \citep{chen2025medium}
& \cmark & \cmark
& \xmark & \xmark
& \cmark & \xmark \\

IF-Beta \citep{wu2026distill}
& \cmark & \xmark
& \xmark & \cmark
& \cmark & \xmark \\

\rowcolor{gray!12}
\textbf{DVA (Ours)}
& \cmark & \cmark
& \cmark & \cmark
& \cmark & \cmark \\

\bottomrule
\end{tabular}%
}
\end{table}
\section{Detailed Experimental Settings}{
\label{sec:app_det_exp_set}
\subsection{Student Training}

To ensure a consistent comparison with prior KD studies, we base our student-training recipes on widely used KD frameworks: RepDistiller~\citep{tian2019crd} for CIFAR-100 and MDistiller~\citep{zhao2023dot} for ImageNet. Unless otherwise specified, we follow their standard optimization and distillation settings, with dataset- and architecture-specific modifications described below. All ResNet experiments on CIFAR-100 are averaged over 10 random seeds, while results in the other experimental settings are averaged over 3 random seeds. All experiments are conducted on NVIDIA GeForce RTX 4090 GPUs.

\paragraph{ResNet on CIFAR-100.}
For the main CIFAR-100 experiments, we use a width-64 ResNet-18 student with $32\times32$ inputs. The student is trained for 240 epochs with a batch size of 64 using SGD with an initial learning rate of $0.05$, momentum of $0.9$, and weight decay of $5\times10^{-4}$. We use a step learning-rate schedule with decay factor $0.1$ at epochs 150, 180, and 210. For knowledge distillation, we use temperature $T=4$ and equally weight the cross-entropy and distillation losses. Unless otherwise specified, the default teacher is a width-64 ResNet-34. Training augmentation consists only of random cropping with four-pixel padding and horizontal flipping. 

\paragraph{ResNet on ImageNet.}
For ImageNet, we use a ResNet-18 student and a pretrained ResNet-34 teacher from \texttt{timm}. The student is trained for 100 epochs with a batch size of 512. We use an initial learning rate of $0.2$, weight decay of $10^{-4}$, and decay the learning rate at epochs 30, 60, and 90. Unlike the CIFAR-100 setting, the distillation temperature is set to $T=1$. Training images are augmented using \texttt{RandomResizedCrop(224)} followed by horizontal flipping.

\paragraph{ResNet on Tiny-ImageNet.}
For Tiny-ImageNet, we retain the same ResNet student-training protocol used for CIFAR-100 except for the input resolution and data augmentation. Images are trained at $64\times64$ resolution using \texttt{RandomResizedCrop(64)} followed by horizontal flipping. At evaluation time, we apply tensor conversion and normalization without additional resizing. For scheduled multi-teacher experiments, we use the width-16 teacher as the auxiliary low-capacity teacher instead of the width-8 teacher used on CIFAR-100 as the dataset size increases.

\paragraph{ViT on CIFAR-100.}
For the ViT experiments on CIFAR-100, we use a ViT-Ti student (\texttt{vit-s4}) and adopt a training recipe tailored to transformer architectures. The student is optimized with AdamW using a learning rate of $2\times10^{-4}$, weight decay of $0.05$, and a cosine learning-rate schedule. We additionally use RandAugment with magnitude 9, random erasing with probability $0.25$, mixup with coefficient $0.8$, and CutMix with coefficient $1.0$.

\subsection{Data Selection Baselines}

For all data-selection baselines, we follow the selection procedures and recommended hyperparameters reported in the original papers and their official implementations, including scoring checkpoints and method-specific sampling parameters. $\mathcal{I}_c$ denotes the indices of samples belonging to class $c$, and $k_c=\lfloor(1-\rho)|\mathcal{I}_c|\rfloor$ denotes the number of samples retained from that class. For methods based on a pruning model, $p_\theta(y_i\mid x_i)$ denotes its predictive probability and $s_i$ denotes the resulting selection score. 

\paragraph{Random.}
Random selection serves as the simplest training-dynamics-free baseline. To match the class-balanced pruning protocol used throughout our experiments, we uniformly sample $k_c$ examples from each class:

\begin{equation}
    \mathcal{M}_c \sim \operatorname{Uniform} \left\{\mathcal{A}\subseteq\mathcal{I}_c: |\mathcal{A}|=k_c \right\}.
\end{equation}

where $\mathcal{A}$ denotes an arbitrary subset of class-$c$ samples with cardinality $k_c$. The final subset is given by $\mathcal{M}=\bigcup_c \mathcal{M}_c$.

\paragraph{Forgetting, GraNd, and EL2N.}
Forgetting~\citep{toneva2019empirical} measures how often a training
example transitions from being correctly classified to incorrectly
classified during optimization.
Let $q_i^{(t)}\in\{0,1\}$ indicate whether sample $i$ is correctly
classified at epoch $j$. Its forgetting score is

\begin{equation}
    s_i^{\mathrm{Forget}}
    =
    \sum_{j=1}^{J-1}
    \mathbb{I}
    \left[
        q_i^{(t)}=1,\,
        q_i^{(t+1)}=0
    \right].
\end{equation}

GraNd and EL2N~\citep{paul2021deep} instead estimate sample importance using gradient and prediction-error signals, respectively:

\begin{equation}
    s_i^{\mathrm{GraNd}} = \left\| \nabla_{\theta} \ell(x_i,y_i;\theta) \right\|_2, \qquad s_i^{\mathrm{EL2N}} = \left\| p_\theta(\cdot\mid x_i)-e_{y_i} \right\|_2,
\end{equation}

where $e_{y_i}$ is the one-hot label vector. Samples with larger scores are preferentially retained according to the original selection procedures. For GraNd, we compute the score at initialization.

\paragraph{CCS.}
Coverage-centric Coreset Selection (CCS) \citep{zheng2023coverage} aims to preserve dataset coverage by stratifying samples according to an importance score rather than retaining only the highest-scoring examples. Following the original implementation, we use the Area Under the Margin (AUM) \citep{pleiss2020identifying} as the importance score. For sample $i$, the margin at epoch $t$ is

\begin{equation}
    m_i^{(t)} = p_{\theta_t}(y_i\mid x_i) - \max_{y\neq y_i}p_{\theta_t}(y\mid x_i),
\end{equation}

and AUM summarizes this margin over training epochs. CCS partitions samples into intervals according to their AUM scores and allocates the pruning budget across these intervals, with a pruning-ratio-dependent cutoff for difficult samples.

\paragraph{D$^2$.}
D$^2$ Pruning~\citep{maharana2024d2} combines sample difficulty and local diversity through message passing over a neighborhood graph. In our experiments, we use the \textbf{Forgetting} score as the initial difficulty score $s_i$, following the corresponding configuration of the original method. Forward message passing updates the score as

\begin{equation}
    \widetilde{s}_i = s_i + \sum_{j\in\mathcal{N}(i)} w_{ij}s_j,
\end{equation}

where $\mathcal{N}(i)$ denotes the neighbors of sample $i$ and $w_{ij}$ their similarity-dependent edge weight. During subset construction, reverse message passing suppresses the scores of samples neighboring already selected examples, thereby discouraging redundancy.

\paragraph{DUAL.}
DUAL~\citep{cho2025lightweight} combines sample difficulty with temporal uncertainty in the ground-truth prediction. For a window of $J$ epochs, let

\begin{equation}
    \bar{p}_i = \frac{1}{J}\sum_{t=1}^{J} p_{\theta_t}(y_i\mid x_i).
\end{equation}

The DUAL score is defined as

\begin{equation}
    s_i^{\mathrm{DUAL}} = (1-\bar{p}_i) \sqrt{ \frac{1}{J-1} \sum_{t=1}^{J} \left( p_{\theta_t}(y_i\mid x_i)-\bar{p}_i \right)^2}.
\end{equation}

Rather than applying a fixed threshold to this score, DUAL further uses a pruning-ratio-adaptive beta sampling policy whose distribution changes with the pruning budget. Samples are drawn without replacement according to the resulting sampling probabilities until the target subset size is reached.

\paragraph{MDSLR.}
MDSLR~\citep{chen2025medium} defines sample difficulty using the cross-entropy loss of the pretrained teacher:

\begin{equation}
    s_i^{\mathrm{MDSLR}} = -\log p_T(y_i\mid x_i).
\end{equation}

Samples are ordered according to this score, and a class-balanced subset from the medium-difficulty region is retained following the original selection rule and hyperparameters.

\paragraph{IF-Beta.}
IF-Beta~\citep{wu2026distill} estimates the utility of each training sample using its influence on a held-out validation set. Let $\mathcal{D}_{\mathrm{val}}$ denote the validation set, $\tilde{g}_i$ the gradient associated with training sample $i$, and $\tilde{H}_{\mathrm{val}}$ the Hessian computed on $\mathcal{D}_{\mathrm{val}}$ at the flat-validation teacher solution. The influence score is defined as

\begin{equation}
    s_i^{\mathrm{IF}} = \tilde{g}_i^\top \tilde{H}_{\mathrm{val}}^{-1} \tilde{g}_i.
\end{equation}

Following the original implementation, the pretrained teacher is first adapted on the validation set to obtain the flat-validation solution used for influence estimation. The resulting scores are rank-normalized and converted into sampling probabilities through a Beta policy. Rather than repeatedly training the full student during policy optimization, IF-Beta constructs a surrogate student by attaching a trainable linear classification head to the frozen teacher feature extractor. The policy location parameter is selected through a discrete grid search, and the remaining policy parameter is optimized using the validation-based bilevel objective. 

\subsection{Teacher Models and Checkpoints}

For the controlled teacher-capacity experiments on CIFAR-100, we construct a family of ResNet-34 teachers by varying the network width while keeping the depth and training recipe fixed. This allows us to change teacher capacity without modifying the overall architecture. Unless otherwise specified, the width-64 ResNet-34 is used as the default large teacher, while the width-8 model is used as the small proxy for data selection. The width-8 proxy requires only about $1/63.3$ of the FLOPs of the width-64 teacher for both training and inference. For the capacity sweep, we use widths $\{4,8,16,32,64,128\}$.

We follow the same principle for our additional controlled experiments, using independently trained teacher models under the same training protocol. For ImageNet, however, we do not train teacher variants specifically for our method. Instead, we directly use publicly available pretrained ResNet checkpoints from \texttt{timm}, including ResNet-10 and ResNet-34, to evaluate whether the proposed selection strategy can operate with off-the-shelf teachers.

\subsection{Teacher Training}

\paragraph{ResNet.}
All teacher variants (except for the ImageNet experiment) are trained independently for 300 epochs using SGD with an initial learning rate of $0.1$, momentum $0.9$, weight decay $5\times10^{-4}$, and a batch size of 128. We use cosine annealing with $T_{\max}=280$ without warmup. Training augmentation consists of random cropping with a padding of 4 and random horizontal flipping, followed by standard CIFAR-100 normalization. All models are trained with automatic mixed precision using a fixed random seed of 42.

\paragraph{ViT.}
For the ViT experiments, training from scratch, we use a scaled ViT-Small architecture with a patch size of $4$ and depth of $12$, and vary the embedding dimension to control model capacity. The models are trained from scratch on CIFAR-100 for 600 epochs using AdamW with an initial learning rate of $10^{-3}$, weight decay of $0.05$, and a batch size of 512. We use cosine learning-rate decay with 40 epochs of warmup, label smoothing of $0.1$, and gradient clipping with a maximum norm of $1.0$. We apply RandAugment ($m=9$, two operations), mixup with $\alpha=0.8$, CutMix with $\alpha=1.0$, and random erasing with probability $0.25$. Training is performed with mixed precision.

For the pretrained ViT experiments, we use \texttt{vit\_small\_patch32\_224.augreg\_in21k} from \texttt{timm}, initialized with ImageNet-21k pretrained weights, and fine-tune it on CIFAR-100. CIFAR-100 images are upsampled to $224\times224$, and the model is fine-tuned for 10,000 optimization steps using SGD with an initial learning rate of $0.01$, a batch size of 512, and no weight decay. We use cosine learning-rate decay with 500 warmup steps and gradient clipping with a maximum norm of $1.0$. Training augmentation consists of random resized cropping with scale $(0.05,1.0)$ and random horizontal flipping, without RandAugment, mixup, CutMix, or random erasing. Fine-tuning is performed with mixed precision.

\subsection{Budget-Aware Difficulty Filtering}
\label{app:diff_filt}

For each class, samples are sorted by the difficulty score in \Cref{eq:difficulty}, and a contiguous interval of the ranking is retained as the candidate pool $\mathcal{P}_c$. \Cref{tab:difficulty_filtering_schedule} reports the exact difficulty region and candidate-pool fraction for each pruning rate. CIFAR-100 and Tiny-ImageNet share the same schedule, whereas ImageNet shifts the filtering region toward harder samples due to its substantially larger number of samples per class. No difficulty filtering is applied at a drop rate of $0.1$.

\begin{table}[H]
\centering
\caption{
Budget-aware difficulty-filtering configurations.
The difficulty region specifies the portion of the class-wise difficulty
ranking retained to construct the candidate pool. The $0.1$ drop rate is omitted because no difficulty filtering is applied.
}
\label{tab:difficulty_filtering_schedule}

\small
\renewcommand{\arraystretch}{1.05}
\setlength{\tabcolsep}{5pt}

\begin{tabular}{c c c c}
\toprule
\multirow{2}{*}{Drop Rate}
& \multicolumn{2}{c}{Difficulty Region}
& \multirow{2}{*}{Pool Size} \\
\cmidrule(lr){2-3}
& CIFAR-100 / Tiny-ImageNet & ImageNet & \\
\midrule
0.3 & Medium & Hard & 80\% \\
0.5 & Medium & Medium & 60\% \\
0.7 & Medium & Medium & 60\% \\
0.8 & Easy & Medium & 40\% \\
0.9 & Easy & Medium-Easy & 20\% \\
\bottomrule
\end{tabular}
\end{table}

}

\newpage
\section{DVA Algorithm}{
\label{app:algo}

\Cref{alg:dva} summarizes the complete DVA selection procedure described in \Cref{sec:4_method}, from budget-aware candidate filtering to class-wise relational projection and greedy volume maximization.

\begin{algorithm}[ht!]
\caption{DVA: Difficulty- and Volume-Aware data selection}
\label{alg:dva}
\begin{algorithmic}[1]

\Require Dataset $\mathcal{S}=\{(x_i,y_i)\}_{i=1}^{N}$,
proxy-teacher outputs $\{p_i\}_{i=1}^{N}$,
drop rate $\rho$, regularization constant $\delta$
\Ensure Selected subset $\mathcal{M}$

\State $\mathcal{M} \gets \emptyset$

\For{each class $c=1,\ldots,C$}

    \State $\mathcal{S}_c \gets \{(x_i,y_i)\in\mathcal{S}:y_i=c\}$
    \State $k_c \gets \lfloor(1-\rho)|\mathcal{S}_c|\rfloor$

    \Statex
    \State \textbf{Stage 1: Difficulty Filtering}

    \State $\mathcal{P}_c
    \gets
    \textsc{DifficultyFiltering}(\mathcal{S}_c,\rho)$
    \Comment{budget-aware candidate pool}

    \Statex
    \State \textbf{Stage 2: Relational Projection}

    \State $\boldsymbol{\mu}_c
    \gets
    \frac{1}{|\mathcal{S}_c|}
    \sum_{(x_i,y_i)\in\mathcal S_c}p_i$

    \State $U_c\Sigma_cV_c^\top
    \gets
    \operatorname{SVD}
    \left(
    [(p_i-\boldsymbol{\mu}_c)^\top]_{(x_i,y_i)\in\mathcal S_c}
    \right)$

    \State $\mathbf{a}_i
    \gets
    \frac{\Sigma_c}{\sqrt{|\mathcal{S}_c|-1}}
    V_c^\top
    (p_i-\boldsymbol{\mu}_c),
    \quad (x_i,y_i)\in\mathcal{P}_c$

    \Statex
    \State \textbf{Stage 3: Volume Maximization}

    \State $\mathcal{M}_c \gets \emptyset$,
    \quad $\mathbf{H} \gets \delta^{-1} I$
    \Comment{$\mathbf{H}=\mathbf{G}^{-1}$}

    \For{$\ell=1,\ldots,k_c$}

        \For{each $(x_i,y_i)\in\mathcal{P}_c\setminus\mathcal{M}_c$}

            \State $s_i
            \gets
            \log\left(
                1+
                \mathbf{a}_i^\top
                \mathbf{H}
                \mathbf{a}_i
            \right)$
            \Comment{marginal volume gain}

        \EndFor

        \State $i^\star
        \gets
        \arg\max_{i:\,(x_i,y_i)\in\mathcal{P}_c\setminus\mathcal{M}_c}
        s_i$

        \State $\mathcal{M}_c
        \gets
        \mathcal{M}_c\cup\{(x_{i^\star},y_{i^\star})\}$

        \State $\mathbf{H}
        \gets
        \mathbf{H}
        -
        \frac{
            \mathbf{H}\mathbf{a}_{i^\star}
            \mathbf{a}_{i^\star}^{\top}\mathbf{H}
        }{
            1+
            \mathbf{a}_{i^\star}^{\top}
            \mathbf{H}
            \mathbf{a}_{i^\star}
        }$
        \Comment{Sherman--Morrison update}


    \EndFor

    \State $\mathcal{M}
    \gets
    \mathcal{M}\cup\mathcal{M}_c$

\EndFor

\State \Return $\mathcal{M}$

\end{algorithmic}
\end{algorithm}

}
\newpage
\section{Additional Analysis of Teacher Supervision}
\label{sec:app_teacher_supervision}

\subsection{Teacher Supervision Decomposition}
\label{sec:app_hybrid_teacher}

\paragraph{Construction of hybrid teachers.}
To isolate the roles of relational ordering and score geometry, we construct hybrid teacher outputs by independently sourcing these two components from different teachers.

For a sample $x$, let $z^A, z^B \in \mathbb{R}^{C}$ denote the logits produced by teachers $A$ and $B$, respectively. We define $\pi(z)$ as the permutation of all classes that sorts the logits in descending order:
\begin{equation}
    z_{\pi(z)_1} \geq z_{\pi(z)_2} \geq \cdots \geq z_{\pi(z)_C}.
\end{equation}
Ties are broken by class index, with the smaller class index ranked first.

We further define the sorted score vector as
\begin{equation} 
    s(z)_r = z_{\pi(z)_r}, \qquad r=1,\ldots,C.
\end{equation}

If a hybrid teacher uses the relational ordering of teacher $A$ and the score geometry of teacher $B$, the hybrid teacher is then defined by assigning the $r$-th largest score of teacher $B$ to the class ranked $r$-th by teacher $A$:
\begin{equation}
    \tilde z^{A_{\mathrm{ord}},B_{\mathrm{geo}}}_{\pi(z^A)_r} = s(z^B)_r, \qquad r=1,\ldots,C.
\label{eq:hybrid_teacher}
\end{equation}

Thus, the hybrid output preserves the complete multiset of scores from teacher $B$, including their magnitudes and margins, while replacing only their assignment to classes according to the ordering induced by teacher $A$. Let $T_L$ and $T_s$ denote the large and small teachers, respectively. When both components are taken from the same teacher, $\tilde z^{T_L{}_{,\mathrm{ord}},T_L{}_{,\mathrm{geo}}}$ and $\tilde z^{T_s{}_{,\mathrm{ord}},T_s{}_{,\mathrm{geo}}}$ reduce to the original outputs of $T_L$ and $T_s$, respectively. The hybrid logits are constructed independently for each sample, after which the same temperature scaling and KD objective as in the original distillation setting are applied.

\paragraph{Additional hybrid teacher results.}

To further examine whether the decomposition observed in the main setting generalizes beyond the original teacher pair, we extend the analysis in three directions. First, on CIFAR-100 with ResNet, we replace the width-8 small teacher with a width-4 and width-16 teachers while keeping the width-64 teacher fixed (\Cref{tab:hybrid_cifar_w4,tab:hybrid_cifar_w16}). Second, we repeat the decomposition on a different dataset, Tiny-ImageNet, using width-64 and width-16 teachers (\Cref{tab:hybrid_tiny_w16}). These additional settings allow us to assess whether the relative contributions of relational ordering and score geometry persist across both teacher capacities and datasets.

On CIFAR-100, we further observe that the relative contribution of the two components varies with teacher capacity. With the width-4 teacher (\Cref{tab:hybrid_cifar_w4}), replacing relational ordering and score geometry individually yields comparable gains of $8.68$ and $8.79$ pp at a drop rate of $0.9$. In the width-8 setting (\Cref{tab:3_rank_geometry_decomposition}), the gain from relational ordering decreases to $3.82$ points, while score geometry still provides an $8.18$-point improvement. This trend becomes more pronounced with the width-16 teacher (\Cref{tab:hybrid_cifar_w16}), where the corresponding gains are $0.50$ and $3.45$ points. These results suggest that relational ordering plays a larger role for lower capacity teachers, whereas the relative contribution of score geometry becomes more dominant as teacher capacity increases.

\begin{table}[H]
\centering
\caption{
Teacher-supervision decomposition on CIFAR-100 with ResNet using width-64 (Large) and width-4 (Small) teachers. Results are averaged over 10 seeds.
}
\label{tab:hybrid_cifar_w4}

\small
\renewcommand{\arraystretch}{1.05}
\setlength{\tabcolsep}{7pt}

\begin{tabular}{llcc}
\toprule
\multirow{2}{*}{Relational Ordering} &
\multirow{2}{*}{Score Geometry} &
\multicolumn{2}{c}{Drop Rate} \\
\cmidrule(lr){3-4}
& & 0.1 & 0.9 \\
\midrule

Large & Large
& 78.82{\scriptsize $\pm$0.25}
& 46.93{\scriptsize $\pm$0.80} \\

Large & Small
& 78.51{\scriptsize $\pm$0.24} {\color{red}$(-0.31)$}
& 55.72{\scriptsize $\pm$0.63}
{\color{green!50!black}$(+8.79)$} \\

Small & Large
& 75.20{\scriptsize $\pm$0.14} {\color{red}$(-3.62)$}
& 55.61{\scriptsize $\pm$0.35}
{\color{green!50!black}$(+8.68)$} \\

Small & Small
& 71.61{\scriptsize $\pm$0.19} {\color{red}$(-7.21)$}
& 59.43{\scriptsize $\pm$0.41}
{\color{green!50!black}$(+12.50)$} \\

\bottomrule
\end{tabular}
\end{table}

\begin{table}[H]
\centering
\caption{
Teacher-supervision decomposition on CIFAR-100 with ResNet using width-64 (Large) and width-16 (Small) teachers. Results are averaged over 10 seeds.
}
\label{tab:hybrid_cifar_w16}

\small
\renewcommand{\arraystretch}{1.05}
\setlength{\tabcolsep}{7pt}

\begin{tabular}{llcc}
\toprule
\multirow{2}{*}{Relational Ordering} &
\multirow{2}{*}{Score Geometry} &
\multicolumn{2}{c}{Drop Rate} \\
\cmidrule(lr){3-4}
& & 0.1 & 0.9 \\
\midrule

Large & Large
& 78.82{\scriptsize $\pm$0.25}
& 46.93{\scriptsize $\pm$0.80} \\

Large & Small
& 78.94{\scriptsize $\pm$0.21} {\color{green!50!black}$(+0.12)$}
& 50.38{\scriptsize $\pm$0.71}
{\color{green!50!black}$(+3.45)$}\\

Small & Large
& 78.69{\scriptsize $\pm$0.25} {\color{red}$(-0.13)$}
& 47.43{\scriptsize $\pm$0.58} {\color{green!50!black}$(+0.50)$} \\

Small & Small
& 78.12 {\scriptsize $\pm$0.15} {\color{red}($-0.70$)}
& 54.10 {\scriptsize $\pm$1.19} {\color{green!50!black}$(+7.17)$} \\

\bottomrule
\end{tabular}
\end{table}

In \Cref{tab:hybrid_tiny_w16}, relational ordering and score geometry contribute comparably under aggressive pruning: replacing only the ordering improves accuracy by $4.48$ pp, while replacing only the score geometry improves accuracy by $4.60$ pp. Thus, the way in which relational ordering contributes varies across settings: it can provide a strong standalone gain or become more consequential when combined with the small teacher's score geometry. Importantly, at a drop rate of $0.9$, the joint configuration performs best in both settings, and removing relational ordering from this configuration consistently reduces accuracy. These results further support preserving relational information when constructing subsets for low-data KD.

\begin{table}[H]
\centering
\caption{
Teacher-supervision decomposition on Tiny-ImageNet with ResNet using width-64 (Large) and width-16 (Small) teachers. Results are averaged over 3 seeds.
}
\label{tab:hybrid_tiny_w16}

\small
\renewcommand{\arraystretch}{1.05}
\setlength{\tabcolsep}{7pt}

\begin{tabular}{llcc}
\toprule
\multirow{2}{*}{Relational Ordering} &
\multirow{2}{*}{Score Geometry} &
\multicolumn{2}{c}{Drop Rate} \\
\cmidrule(lr){3-4}
& & 0.1 & 0.9 \\
\midrule

Large & Large
& 66.20{\scriptsize $\pm$0.47}
& 45.03{\scriptsize $\pm$0.64} \\

Large & Small
& 66.49{\scriptsize $\pm$0.16} {\color{green!50!black}$(+0.29)$}
& 49.63{\scriptsize $\pm$0.41} {\color{green!50!black}$(+4.60)$} \\

Small & Large
& 66.60{\scriptsize $\pm$0.15} {\color{green!50!black}$(+0.40)$}
& 49.51{\scriptsize $\pm$0.26} {\color{green!50!black}$(+4.48)$} \\

Small & Small
& 65.62{\scriptsize $\pm$0.38} {\color{red}$(-0.58)$}
& 52.56{\scriptsize $\pm$0.37} {\color{green!50!black}$(+7.53)$} \\

\bottomrule
\end{tabular}
\end{table}

\subsection{KD Performance Across Teacher Capacities}{
\label{sec:app_teacher_sweep_kd}

We further examine how the preferred teacher capacity changes with the available data budget across different datasets and architectures. Across the ResNet-based experiments (\Cref{tab:app_res_c100_teacher_cap_kd,tab:app_tinyimagenet_teacher_width,tab:app_imagenet_teacher_depth}), a common trend emerges: as the data budget decreases, students distilled from lower-capacity teachers tend to achieve better performance. On CIFAR-100 (\Cref{tab:app_res_c100_teacher_cap_kd}), the preferred teacher gradually shifts from wider teachers at larger data budgets to narrower teachers under aggressive pruning. A similar trend appears on Tiny-ImageNet (\Cref{tab:app_tinyimagenet_teacher_width}), where the width-32 teacher performs best over most pruning rates, while the width-16 teacher becomes preferable at the most severe pruning rate.


\begin{table}[H]
\centering
\caption{KD accuracy (\%) on CIFAR-100 with a ResNet-18 student and
ResNet-34 teachers of varying widths. Results are averaged over 10 seeds. Best/second-best results are in \textbf{bold}/\underline{underline}.}
\label{tab:app_res_c100_teacher_cap_kd}

\resizebox{\linewidth}{!}{%
\begin{tabular}{lccccccccc}
\toprule
\multirow{2}{*}{Teacher W.} &
\multirow{2}{*}{T. Train Acc.} &
\multirow{2}{*}{T. Val. Acc.} &
\multicolumn{7}{c}{Drop Rate} \\
\cmidrule{4-10}
& & & 0.0 & 0.1 & 0.3 & 0.5 & 0.7 & 0.8 & 0.9 \\
\midrule

128
& 99.98
& \textbf{80.95}
& 79.35
& 78.49
& 75.87
& 72.50
& 66.20
& 60.82
& 44.81 \\

64 \textit{(Default)}
& 99.97
& {\ul 79.71}
& {\ul 79.73}
& \textbf{78.82}
& 76.69
& 73.16
& 67.07
& 61.35
& 46.93 \\

32
& 99.98
& 77.87
& \textbf{79.80}
& {\ul 78.80}
& {\ul 77.16}
& {\ul 74.40}
& 68.88
& 63.81
& 47.84 \\

16
& 99.30
& 73.17
& 78.65
& 78.12
& \textbf{77.17}
& \textbf{75.08}
& \textbf{71.31}
& {\ul 67.49}
& 54.10 \\

8
& 91.87
& 67.44
& 75.42
& 75.33
& 74.64
& 73.54
& {\ul 71.24}
& \textbf{69.09}
& \textbf{62.49} \\

4
& 65.57
& 57.28
& 72.14
& 71.70
& 70.65
& 68.53
& 65.79
& 63.78
& {\ul 59.33} \\

\bottomrule
\end{tabular}%
}
\end{table}

\begin{table}[H]
\centering
\caption{
KD accuracy (\%) on Tiny-ImageNet with a ResNet-18 student and ResNet-34 teachers of varying widths. Results are averaged over 3 seeds. Best/second-best results are in \textbf{bold}/\underline{underline}.
}
\label{tab:app_tinyimagenet_teacher_width}
\renewcommand{\arraystretch}{1.15}

\resizebox{\linewidth}{!}{
\begin{tabular}{l cc ccccccc}
\toprule
\multirow{2}{*}{Teacher Width} &
\multirow{2}{*}{T. Train Acc.} &
\multirow{2}{*}{T. Val. Acc.} &
\multicolumn{6}{c}{Drop Rate} \\
\cmidrule(lr){4-10}
& & & 0.0 & 0.1 & 0.3 & 0.5 & 0.7 & 0.8 & 0.9 \\[1pt]
\midrule

64 \textit{(Default)}
& 99.38 & \textbf{65.57}
& {\ul 66.29} & {\ul 66.20} & {\ul 64.45} & 62.23 & 57.45 & 54.11 & 45.03 \\

32
& 91.69 & {\ul 63.26}
& \textbf{66.86} & \textbf{67.31} & \textbf{66.19} & \textbf{64.46}
& \textbf{61.38} & \textbf{58.04} & {\ul 50.66} \\

16
& 71.81 & 57.54
& 65.65 & 65.62 & 64.36 & {\ul 63.00} & {\ul 60.47} & {\ul 57.87} & \textbf{52.56} \\

8
& 54.61 & 49.32
& 62.94 & 62.68 & 61.37 & 59.41 & 56.25 & 53.78 & 49.39 \\

\bottomrule
\end{tabular}
}
\end{table}

Importantly, this behavior is not limited to our controlled width-scaling experiments. On ImageNet (\Cref{tab:app_imagenet_teacher_depth}), where we use publicly available pretrained ResNet teachers with different depths, the preferred teacher similarly shifts toward shallower models as the data budget decreases. ResNet-34 performs best at drop rates of $0.1$, $0.3$ and $0.5$, whereas ResNet-18 and ResNet-10 become preferable at drop rates of $0.7$ and $0.8$ and $0.9$, respectively. This suggests that the budget-dependent shift in teacher preference is not specific to width scaling or to teachers trained under a controlled experimental setup.

\begin{table}[H]
\centering
\caption{
KD accuracy (\%) on ImageNet with a ResNet-18 student and teachers of varying depths. Results are averaged over 3 seeds.
Best/second-best results are in \textbf{bold}/\underline{underline}.
}
\label{tab:app_imagenet_teacher_depth}

\small
\renewcommand{\arraystretch}{1.12}
\setlength{\tabcolsep}{4.5pt}
\resizebox{\linewidth}{!}{
\begin{tabular}{l cc ccccccc}
\toprule
\multirow{2}{*}{Teacher}
& \multirow{2}{*}{T. Train Acc.}
& \multirow{2}{*}{T. Val. Acc.}
& \multicolumn{7}{c}{Drop Rate} \\
\cmidrule(lr){4-10}
& & & 0.0 & 0.1 & 0.3 & 0.5 & 0.7 & 0.8 & 0.9 \\
\midrule

ResNet-50
& \textbf{92.93}
& \textbf{80.11}
& \underline{71.43}
& \underline{71.08}
& 69.85
& 67.75
& 63.18
& 58.40
& 49.03 \\

ResNet-34
& \underline{85.06}
& \underline{76.32}
& \textbf{71.68}
& \textbf{71.32}
& \textbf{70.24}
& \textbf{68.49}
& 64.31
& 59.96
& 50.75 \\

ResNet-18
& 78.62
& 71.47
& \underline{71.43}
& 71.01
& \underline{70.05}
& \underline{68.45}
& \textbf{64.99}
& \underline{61.05}
& \underline{52.66} \\

ResNet-10
& 75.77
& 68.23
& 70.73
& 70.38
& 69.48
& 68.03
& \underline{64.85}
& \textbf{61.40}
& \textbf{54.04} \\

\bottomrule
\end{tabular}
}%
\end{table}

The ViT experiments (\Cref{tab:cifar100_vit_teacher_width}) exhibit a less monotonic relationship between teacher capacity and student performance, with a roughly U-shaped trend across embedding dimensions. Nevertheless, the same qualitative behavior remains visible in the low-data regime. At a $0.9$ drop rate, the teacher with embedding dimension 48 yields $43.68\%$ student accuracy, outperforming both the ImageNet-pretrained teacher with embedding dimension 384 ($41.66\%$) and the corresponding from-scratch teacher ($37.79\%$). Thus, while the optimal teacher capacity does not necessarily decrease monotonically for every architecture, lower-capacity teachers can become substantially more effective as the available training data becomes limited.

\begin{table}[H]
\centering
\caption{
KD accuracy (\%) on CIFAR-100 with a ViT-Ti student and ViT-S teachers of varying widths. Results are averaged over 3 seeds. Best/second-best results are in \textbf{bold}/\underline{underline}.
}
\label{tab:cifar100_vit_teacher_width}
\renewcommand{\arraystretch}{1.15}

\resizebox{\linewidth}{!}{
\begin{tabular}{l cc ccccccc}
\toprule
\multirow{2}{*}{Teacher Embedding} &
\multirow{2}{*}{T. Train Acc.} &
\multirow{2}{*}{T. Val. Acc.} &
\multicolumn{7}{c}{Drop Rate} \\
\cmidrule(lr){4-10}
& & & 0.0 & 0.1 & 0.3 & 0.5 & 0.7 & 0.8 & 0.9 \\[1pt]
\midrule

384 \textit{(Pre-trained)}
& 99.35
& \textbf{90.53}
& \textbf{75.38}
& \textbf{73.79}
& \textbf{70.95}
& \textbf{66.75}
& \textbf{60.11}
& \textbf{54.28}
& {\ul 41.66} \\

384 \textit{(From Scratch)}
& \textbf{99.97}
& 68.58
& 70.41
& 69.31
& 66.08
& 61.64
& 55.13
& 49.59
& 37.79 \\

192
& {\ul 99.96}
& {\ul 68.81}
& 70.82
& 68.85
& 64.61
& 57.75
& 49.13
& 40.39
& 28.44 \\

96
& 95.06
& 65.74
& {\ul 72.69}
& {\ul 70.66}
& 66.95
& 60.34
& 51.70
& 43.54
& 30.45 \\

48
& 65.85
& 49.76
& 71.06
& 69.84
& {\ul 67.64}
& {\ul 64.38}
& {\ul 58.06}
& {\ul 54.01}
& \textbf{43.68} \\

\bottomrule
\end{tabular}
}
\end{table}

}
\subsection{Temperature Scaling Analysis}
\label{sec:app_temperature}

One possible explanation for the low-data advantage of smaller teachers is that
the default distillation temperature may be suboptimal for larger teachers.
We therefore sweep the KD temperature from $1$ to $32$ for teachers of different
capacities under $90\%$ random pruning.

As shown in \Cref{tab:app_temperature_sweep}, temperature tuning improves the performance of larger teachers but does not close the gap to the smaller teacher. The width-64 teacher achieves its best accuracy of $47.29\%$ at $T=8$, whereas the width-8 teacher reaches $63.69\%$ at its best temperature, leaving a $16.40$ pp gap even after tuning temperature separately for each teacher. The gap is also similar to that observed at the default $T=4$ ($16.74$ pp; cf. 15.56 pp in Table 1, which uses 10 seeds)), indicating that the budget-dependent teacher preference is not an artifact of a particular temperature choice.

Temperature scaling also provides an ordering-preserving intervention on teacher predictions: changing $T$ alters the score geometry while leaving the class ordering unchanged. The persistence of the teacher gap across temperatures therefore provides additional evidence that adjusting score geometry alone, at least within the family induced by temperature scaling, cannot fully explain the low-data transfer gap. This observation is consistent with our decomposition analysis, where relational ordering provides an additional source of useful supervision.

\begin{table}[H]
\centering
\caption{
KD accuracy (\%) under different distillation temperatures on CIFAR-100
with $90\%$ random pruning.
Results are averaged over 3 seeds and reported as mean $\pm$ standard deviation.
Temperature is varied independently for each teacher while all other settings are fixed.
Best results in each row are in \textbf{bold}.}
\label{tab:app_temperature_sweep}

\small
\renewcommand{\arraystretch}{1.08}
\setlength{\tabcolsep}{5pt}

\begin{tabular}{lcccccc}
\toprule
\multirow{2}{*}{Teacher}
& \multicolumn{6}{c}{KD Temperature} \\
\cmidrule(lr){2-7}
& $T{=}1$ & $T{=}2$ & $T{=}4$ & $T{=}8$ & $T{=}16$ & $T{=}32$ \\
\midrule

Width-4
& 56.74{\scriptsize $\pm$0.35}
& 59.02{\scriptsize $\pm$0.24}
& 59.44{\scriptsize $\pm$0.39}
& 59.30{\scriptsize $\pm$0.30}
& \textbf{59.66}{\scriptsize $\pm$0.23}
& 59.45{\scriptsize $\pm$0.14} \\

Width-8
& 49.14{\scriptsize $\pm$0.98}
& 57.68{\scriptsize $\pm$0.73}
& 62.16{\scriptsize $\pm$0.36}
& \textbf{63.69}{\scriptsize $\pm$0.18}
& 63.51{\scriptsize $\pm$0.21}
& 63.60{\scriptsize $\pm$0.28} \\

Width-16
& 41.16{\scriptsize $\pm$1.00}
& 45.30{\scriptsize $\pm$1.78}
& 53.92{\scriptsize $\pm$1.26}
& 56.94{\scriptsize $\pm$1.15}
& 57.91{\scriptsize $\pm$1.18}
& \textbf{58.30}{\scriptsize $\pm$1.28} \\

Width-32
& 39.66{\scriptsize $\pm$1.67}
& 42.57{\scriptsize $\pm$1.08}
& 48.00{\scriptsize $\pm$1.24}
& 49.32{\scriptsize $\pm$1.45}
& 50.52{\scriptsize $\pm$1.03}
& \textbf{50.54}{\scriptsize $\pm$0.33} \\

Width-64
& 39.21{\scriptsize $\pm$1.02}
& 41.19{\scriptsize $\pm$1.52}
& 45.42{\scriptsize $\pm$1.70}
& \textbf{47.29}{\scriptsize $\pm$1.04}
& 46.57{\scriptsize $\pm$2.20}
& 46.61{\scriptsize $\pm$0.41} \\

\bottomrule
\end{tabular}
\end{table}
\section{Additional Analysis of DVA}

\subsection{Interpretation of Principal Relational Directions}
\label{sec:app_principal_directions}

To better understand the relational projection used in \Cref{sec:4_method}, we examine what the principal directions obtained from the class-conditional teacher outputs represent.
Recall from \Cref{eq:class_svd} that we decompose the centered class-conditional output matrix as $X_c = U_c \Sigma_c V_c^\top$. The columns $\{v_r\}_{r=1}^{R}$ of $V_c$ define the principal directions of variation in the teacher outputs, where each entry $v_r[j]$ measures the contribution of class dimension $j$ to direction $v_r$. We refer to $|v_r[j]|$ as its \emph{direction weight}.

\begin{figure}[H]
    \centering

    \begin{subfigure}[t]{0.31\textwidth}
        \centering
        \includegraphics[width=\linewidth]{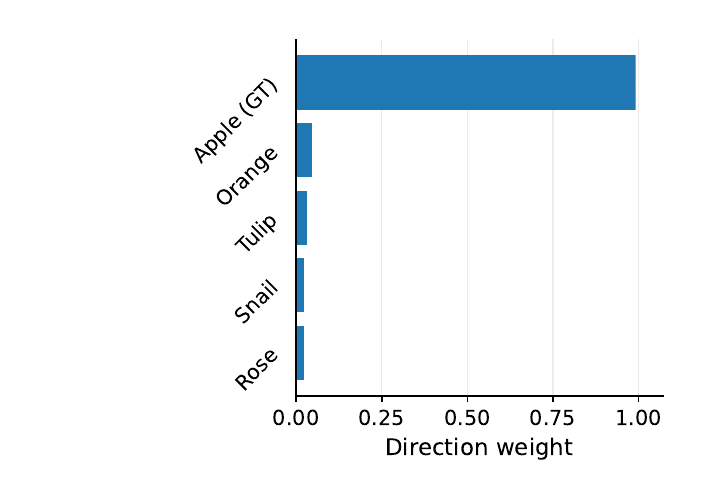}
        \caption{\textit{apple}, $v_{1}$ (78.2\%)}
    \end{subfigure}
    \hfill
    \begin{subfigure}[t]{0.31\textwidth}
        \centering
        \includegraphics[width=\linewidth]{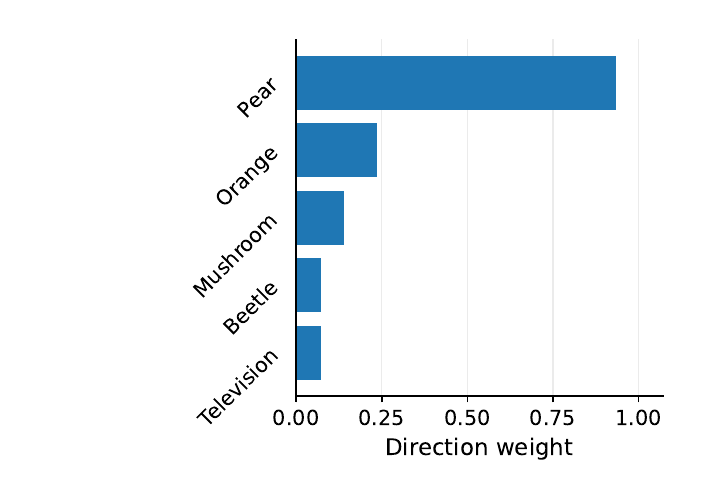}
        \caption{\textit{apple}, $v_{2}$ (6.7\%)}
    \end{subfigure}
    \hfill
    \begin{subfigure}[t]{0.31\textwidth}
        \centering
        \includegraphics[width=\linewidth]{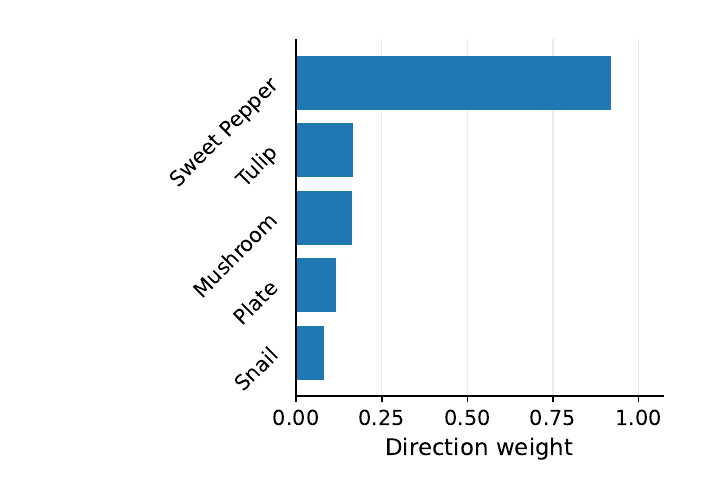}
        \caption{\textit{apple}, $v_{3}$ (4.5\%)}
    \end{subfigure}

    \begin{subfigure}[t]{0.31\textwidth}
        \centering
        \includegraphics[width=\linewidth]{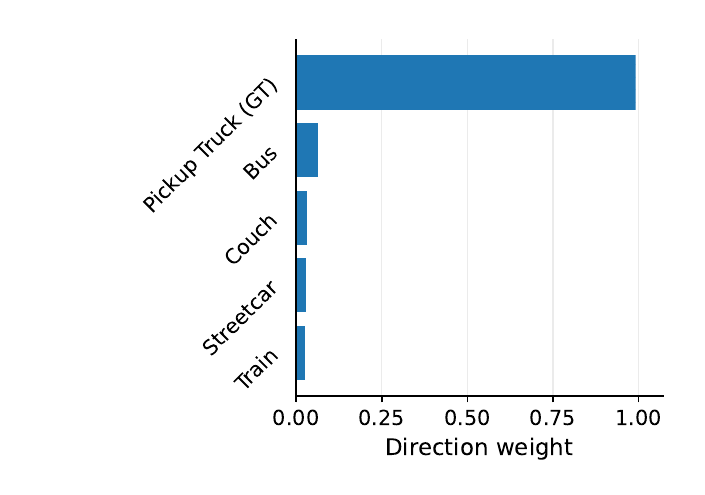}
        \caption{\textit{pickup truck}, $v_{1}$ (80.6\%)}
    \end{subfigure}
    \hfill
    \begin{subfigure}[t]{0.31\textwidth}
        \centering
        \includegraphics[width=\linewidth]{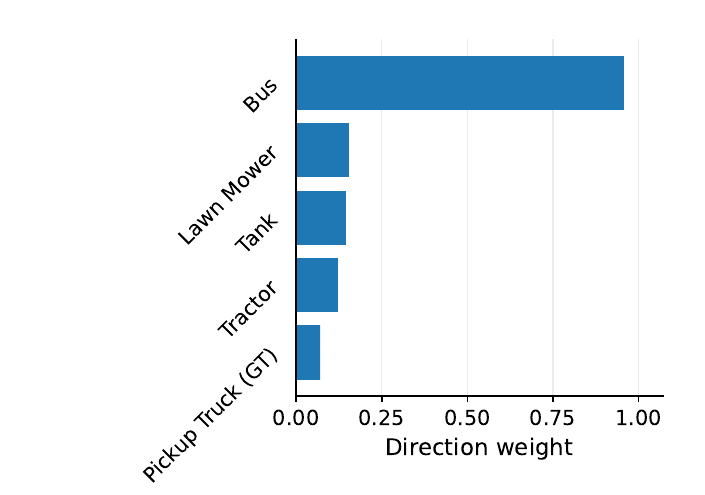}
        \caption{\textit{pickup truck}, $v_{2}$ (7.6\%)}
    \end{subfigure}
    \hfill
    \begin{subfigure}[t]{0.31\textwidth}
        \centering
        \includegraphics[width=\linewidth]{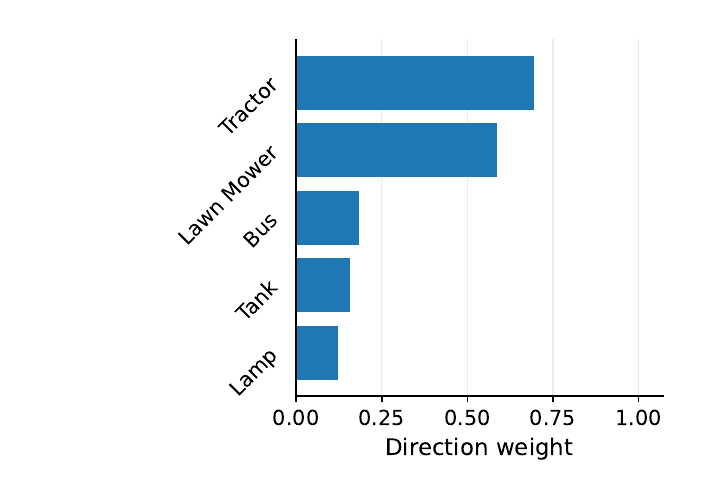}
        \caption{\textit{pickup truck}, $v_{3}$ (3.1\%)}
    \end{subfigure}

    \begin{subfigure}[t]{0.31\textwidth}
        \centering
        \includegraphics[width=\linewidth]{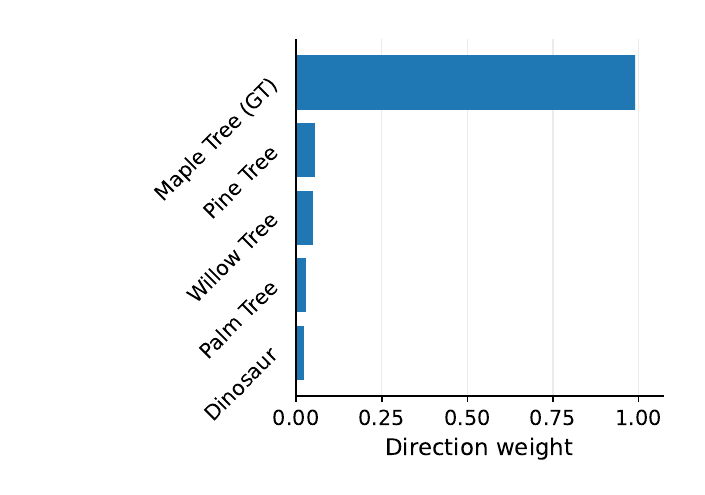}
        \caption{\textit{maple tree}, $v_{1}$ (62.6\%)}
    \end{subfigure}
    \hfill
    \begin{subfigure}[t]{0.31\textwidth}
        \centering
        \includegraphics[width=\linewidth]{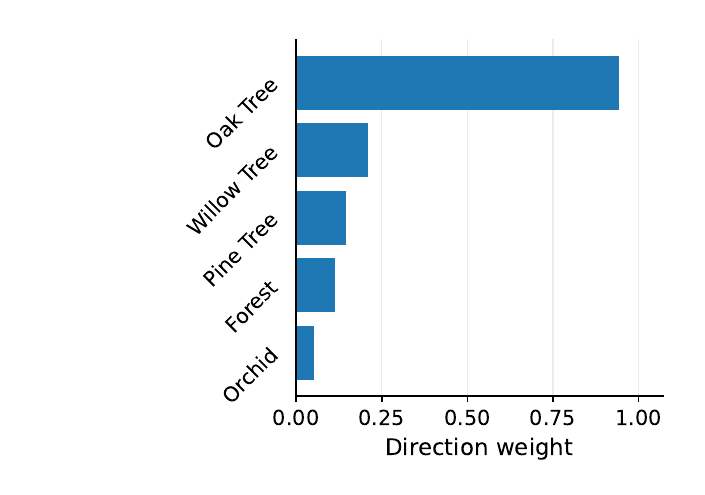}
        \caption{\textit{maple tree}, $v_{2}$ (18.8\%)}
    \end{subfigure}
    \hfill
    \begin{subfigure}[t]{0.31\textwidth}
        \centering
        \includegraphics[width=\linewidth]{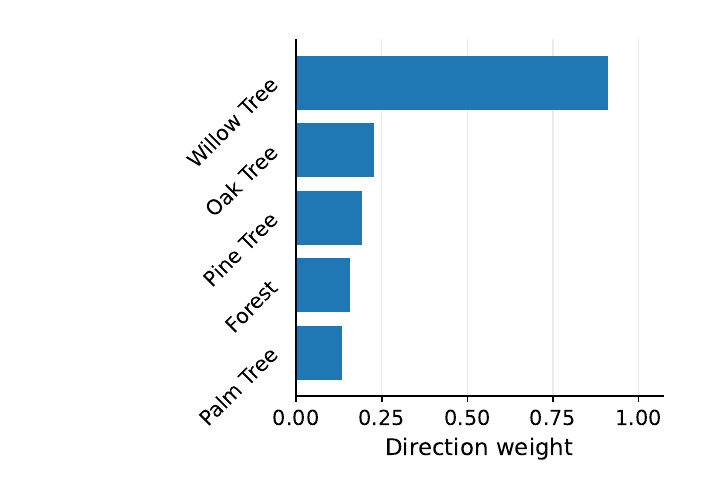}
        \caption{\textit{maple tree}, $v_{3}$ (8.0\%)}
    \end{subfigure}

    \caption{ \textbf{Principal directions of class-conditional outputs from a ResNet-34 width-8 teacher on CIFAR-100.} For each ground-truth class, we center the teacher output vectors over samples in that class and perform SVD on the resulting matrix. Each row shows the first three principal directions, $v_1$, $v_2$, and $v_3$, for one representative class. Within each subfigure, the bars show the five class dimensions with the largest absolute entries in the corresponding direction, referred to as \emph{direction weights}. The percentage in each subcaption denotes the fraction of class-conditional variance explained by that direction. Hatched bars indicate the ground-truth (GT) class.}
    \label{fig:principal_directions_examples}
\end{figure}

As shown in \Cref{fig:principal_directions_examples}, the leading direction $v_1$ is dominated by the ground-truth class, indicating that the largest source of variation is associated with the target response. In contrast, subsequent directions are primarily composed of non-target classes that are semantically related to, or frequently confused with, the ground-truth class. For example, the dominant non-target dimensions include pear and sweet pepper for \textit{apple}, bus and tractor for \textit{pickup truck}, and oak tree and willow tree for \textit{maple tree}. These patterns indicate that the lower-order principal directions capture recurring class confusion structure in the teacher outputs rather than arbitrary logit variation.

This provides an interpretation for the relational projection in \Cref{sec:4_method}: projecting samples onto these directions emphasizes class-conditional variations associated with the teacher's non-target relations. The subsequent volume-based selection can therefore favor subsets that span complementary relational directions, rather than repeatedly selecting samples that express similar teacher outputs.

\subsection{Effect of Volume Maximization}
\label{sec:app_volume_search}

As an extension of the analysis in \Cref{sec:5_effect}, we examine how much of the performance gain comes from the volume maximization step itself. To isolate this effect from difficulty filtering, we keep the filtered candidate pool fixed and compare three selection strategies: random sampling, our volume maximizing greedy selection (highest volume), and a reverse greedy variant that selects the sample with the smallest marginal volume gain at each step (lowest volume). As shown in \Cref{tab:res_vol}, volume maximization consistently outperforms filtered random across all data budgets, whereas minimizing the marginal volume gain consistently degrades KD performance. The difference becomes particularly pronounced under aggressive pruning: at drop rates of $0.7$, $0.8$, and $0.9$, the reverse greedy subsets fall below Filtered Random by $2.95$, $3.01$, and $2.44$ pp, respectively, while volume maximization improves accuracy by $1.00$, $1.05$, and $0.74$ pp. These results show that the gain is not explained by difficulty filtering alone, but also depends on selecting subsets with complementary relational supervision.

\begin{table}[H]

\centering
\caption{
Effect of relational volume on KD accuracy across data budgets. All methods select from the same difficulty-filtered candidate pool. Results are averaged over 10 seeds. Best results are in \textbf{bold}.
}
\label{tab:res_vol}

\footnotesize
\renewcommand{\arraystretch}{1.08}
\setlength{\tabcolsep}{3.5pt}

\begin{tabular}{lcccccc}
\toprule
& \multicolumn{6}{c}{Drop Rate} \\
\cmidrule(lr){2-7}
Selection Strategy
& 0.1 & 0.3 & 0.5 & 0.7 & 0.8 & 0.9 \\
\midrule

Filtered Random
& 78.78{\scriptsize$\pm$0.20}
& 76.78{\scriptsize$\pm$0.20}
& 74.31{\scriptsize$\pm$0.31}
& 68.09{\scriptsize$\pm$0.20}
& 63.40{\scriptsize$\pm$0.32}
& 53.02{\scriptsize$\pm$0.62} \\

Lowest Volume
& 78.44{\scriptsize$\pm$0.35}
& 76.51{\scriptsize$\pm$0.17}
& 73.74{\scriptsize$\pm$0.19}
& 65.14{\scriptsize$\pm$0.36}
& 60.39{\scriptsize$\pm$0.27}
& 50.58{\scriptsize$\pm$0.51} \\

Highest Volume (Ours)
& \textbf{79.13}{\scriptsize$\pm$0.17}
& \textbf{77.43}{\scriptsize$\pm$0.20}
& \textbf{74.62}{\scriptsize$\pm$0.16}
& \textbf{69.09}{\scriptsize$\pm$0.23}
& \textbf{64.45}{\scriptsize$\pm$0.36}
& \textbf{53.76}{\scriptsize$\pm$0.77} \\

\bottomrule
\end{tabular}

\end{table}

\subsection{Sensitivity to the Regularization Parameter}

We next examine the sensitivity of DVA to the regularization parameter $\delta$ (\Cref{eq:logdet_objective}). As shown in \Cref{tab:app_delta_sweep}, DVA is fairly stable across small values of $\delta$. In particular, for $\delta \in \{10^{-6},10^{-5},10^{-4},10^{-3},10^{-2}\}$, the overall performance varies only modestly across pruning rates. By contrast, when $\delta$ becomes large, performance degrades more noticeably, especially under aggressive pruning. For example, at a drop rate of $0.9$, accuracy decreases from the range of $53.47$--$54.19\%$ for small $\delta$ values to $52.72\%$ at $\delta=1$ and $51.62\%$ at $\delta=10$. These results indicate that DVA is robust to the choice of $\delta$ as long as $\delta$ remains sufficiently small, and validate our default choice of $\delta=10^{-3}$.

\begin{table}[H]
\centering
\caption{
Effect of the regularization parameter $\delta$ in DVA on CIFAR-100.
Results are averaged over 10 seeds.
}
\label{tab:app_delta_sweep}

\small
\renewcommand{\arraystretch}{1.12}
\setlength{\tabcolsep}{5pt}

\begin{tabular}{lcccccc}
\toprule
& \multicolumn{6}{c}{Drop Rate} \\
\cmidrule(lr){2-7}
$\delta$
& 0.1 & 0.3 & 0.5 & 0.7 & 0.8 & 0.9 \\
\midrule

$10^{-6}$
& 79.05{\scriptsize $\pm$0.25}
& 77.26{\scriptsize $\pm$0.15}
& 74.60{\scriptsize $\pm$0.19}
& 69.17{\scriptsize $\pm$0.27}
& 64.72{\scriptsize $\pm$0.37}
& 53.82{\scriptsize $\pm$0.67} \\

$10^{-5}$
& 79.02{\scriptsize $\pm$0.29}
& 77.39{\scriptsize $\pm$0.20}
& 74.72{\scriptsize $\pm$0.22}
& 68.95{\scriptsize $\pm$0.28}
& 64.62{\scriptsize $\pm$0.22}
& 54.19{\scriptsize $\pm$0.73} \\

$10^{-4}$
& 79.05{\scriptsize $\pm$0.17}
& 77.37{\scriptsize $\pm$0.19}
& 74.66{\scriptsize $\pm$0.30}
& 69.12{\scriptsize $\pm$0.17}
& 64.46{\scriptsize $\pm$0.47}
& 53.74{\scriptsize $\pm$0.56} \\

\rowcolor{gray!12}
$10^{-3}$ (default)
& 79.13{\scriptsize $\pm$0.17}
& 77.43{\scriptsize $\pm$0.20}
& 74.62{\scriptsize $\pm$0.16}
& 69.09{\scriptsize $\pm$0.23}
& 64.45{\scriptsize $\pm$0.36}
& 53.76{\scriptsize $\pm$0.77} \\

$10^{-2}$
& 79.04{\scriptsize $\pm$0.27}
& 77.56{\scriptsize $\pm$0.28}
& 74.73{\scriptsize $\pm$0.21}
& 68.85{\scriptsize $\pm$0.30}
& 64.42{\scriptsize $\pm$0.26}
& 53.47{\scriptsize $\pm$0.53} \\

$1$
& 78.85{\scriptsize $\pm$0.26}
& 77.34{\scriptsize $\pm$0.17}
& 74.60{\scriptsize $\pm$0.19}
& 68.27{\scriptsize $\pm$0.28}
& 63.26{\scriptsize $\pm$0.35}
& 52.72{\scriptsize $\pm$0.66} \\

$10$
& 79.02{\scriptsize $\pm$0.13}
& 77.30{\scriptsize $\pm$0.22}
& 74.29{\scriptsize $\pm$0.52}
& 67.54{\scriptsize $\pm$0.10}
& 62.55{\scriptsize $\pm$0.20}
& 51.62{\scriptsize $\pm$0.32} \\

\bottomrule
\end{tabular}
\end{table}
\subsection{Proxy Teacher Ablation}

\paragraph{Early-checkpoint proxy.} In this section, we ask whether the advantage of the small proxy can be reproduced by an earlier checkpoint of the large teacher. We therefore use width-64 checkpoints from epochs 10, 20, 30, and 40 as the DVA proxy, while keeping the final distillation teacher fixed to the fully trained width-64 model. As shown in \Cref{tab:app_early_proxy}, early checkpoints recover most of the large proxy degradation under aggressive pruning. At a drop rate of $0.9$, accuracy increases from $35.96\%$ with the fully trained width-64 proxy to $53.47\%$ with the best early checkpoint, which is comparable to the $53.76\%$ obtained with the width-8 proxy. The width-8 proxy remains clearly stronger at some budgets, including drop rates of $0.7$ and $0.8$, where it achieves $69.09\%$ and $64.45\%$ compared with $68.70\%$ and $63.38\%$ for the best early width-64 checkpoints.

These results show that the training stage of the proxy explains the performance gap between fully trained large and small proxies. Using an early checkpoint of the width-64 proxy substantially improves performance over its fully trained counterpart, particularly at high pruning rates. However, no single early checkpoint performs best across all budgets, requiring checkpoint selection for each pruning rate. In contrast, the fully trained width-8 proxy matches or outperforms the best early width-64 checkpoint at five of the six pruning rates, while requiring no checkpoint search and being cheaper to use as a selection proxy due to having approximately $63.3\times$ fewer parameters than the width-64 proxy.

\begin{table}[H]
\centering
\caption{
Effect of using early checkpoints of the width-64 ResNet-34 teacher (Large) as the DVA proxy on CIFAR-100. DVA (Ours) uses the width-8 proxy (Small), and all runs are averaged over 10 seeds. Best results are in \textbf{bold}.
}
\label{tab:app_early_proxy}
\resizebox{0.95\linewidth}{!}{
\begin{tabular}{lcccccc}
\toprule
& \multicolumn{6}{c}{Drop Rate} \\
\cmidrule(lr){2-7}
Proxy 
& 0.1 & 0.3 & 0.5 & 0.7 & 0.8 & 0.9 \\
\midrule

\rowcolor{gray!12}
Small (DVA, Ours) 
& 79.13{\scriptsize $\pm$0.17}
& \textbf{77.43}{\scriptsize $\pm$0.20}
& \textbf{74.62}{\scriptsize $\pm$0.16}
& \textbf{69.09}{\scriptsize $\pm$0.23}
& \textbf{64.45}{\scriptsize $\pm$0.36}
& \textbf{53.76}{\scriptsize $\pm$0.77} \\

\midrule

Large @ ep10 
& 79.02{\scriptsize $\pm$0.24}
& 76.86{\scriptsize $\pm$0.24}
& 73.53{\scriptsize $\pm$0.26}
& 67.27{\scriptsize $\pm$0.23}
& 62.28{\scriptsize $\pm$0.25}
& 52.47{\scriptsize $\pm$0.44} \\

Large @ ep20 
& 79.05{\scriptsize $\pm$0.14}
& 77.38{\scriptsize $\pm$0.26}
& 74.24{\scriptsize $\pm$0.24}
& 68.64{\scriptsize $\pm$0.15}
& 63.14{\scriptsize $\pm$0.19}
& 53.47{\scriptsize $\pm$0.54} \\

Large @ ep30 
& \textbf{79.26}{\scriptsize $\pm$0.28}
& 77.14{\scriptsize $\pm$0.19}
& 74.24{\scriptsize $\pm$0.15}
& 68.70{\scriptsize $\pm$0.31}
& 63.33{\scriptsize $\pm$0.20}
& 52.63{\scriptsize $\pm$0.56} \\

Large @ ep40 
& 79.08{\scriptsize $\pm$0.22}
& 77.34{\scriptsize $\pm$0.14}
& 74.21{\scriptsize $\pm$0.18}
& 68.44{\scriptsize $\pm$0.28}
& 63.38{\scriptsize $\pm$0.29}
& 52.11{\scriptsize $\pm$0.32} \\

Large @ final
& 79.12{\scriptsize$\pm$0.28}
& 76.62{\scriptsize$\pm$0.20}
& 73.25{\scriptsize$\pm$0.45}
& 64.09{\scriptsize$\pm$0.25}
& 55.45{\scriptsize$\pm$0.63}
& 35.96{\scriptsize$\pm$1.14} \\

\bottomrule
\end{tabular}
}
\end{table}

\paragraph{Stage-wise proxy ablation.} \Cref{tab:app_proxy_ablation} disentangles the roles of the proxy in difficulty filtering and volume maximization. The choice of proxy is particularly important for constructing the candidate pool: at a drop rate of $0.9$, using the small teacher for difficulty filtering yields $53.76\%$, whereas replacing it with the large teacher reduces accuracy to $39.94\%$, even when the small teacher is retained for volume maximization. Once a suitable candidate pool is formed, however, the volume maximization stage is robust to the proxy choice, with the small and large proxies giving nearly identical performance ($53.76\%$ vs. $53.74\%$). Importantly, volume maximization still provides consistent gains over random selection within the same filtered pool across budgets, including a $1.05$-point improvement at a drop rate of $0.8$ ($64.45\%$ vs. $63.40\%$). These results highlight complementary roles of the two stages: difficulty filtering determines which samples are suitable for the available budget, while volume maximization further selects a complementary subset from this candidate pool.

\begin{table}[H]
\centering
\caption{
Effect of the proxy used for difficulty filtering and volume maximization on CIFAR-100.
Small and Large denote the width-8 and width-64 teachers, and all runs are averaged over 10 seeds. Best results are in \textbf{bold}.
}
\label{tab:app_proxy_ablation}

\small
\renewcommand{\arraystretch}{1.05}
\setlength{\tabcolsep}{4.5pt}

\begin{tabular}{llcccccc}
\toprule
\multirow{2}{*}{\shortstack[c]{Difficulty\\[-1.5pt]Filtering}}
& \multirow{2}{*}{\shortstack[c]{Volume\\[-1.5pt]Maximization}}
& \multicolumn{6}{c}{Drop Rate} \\
\cmidrule(lr){3-8}
& & 0.1 & 0.3 & 0.5 & 0.7 & 0.8 & 0.9 \\
\midrule

\rowcolor{gray!12}
Small & Small
& \textbf{79.13}{\scriptsize $\pm 0.17$}
& \textbf{77.43}{\scriptsize $\pm 0.20$}
& \textbf{74.62}{\scriptsize $\pm 0.16$}
& \textbf{69.09}{\scriptsize $\pm 0.23$}
& \textbf{64.45}{\scriptsize $\pm 0.36$}
& \textbf{53.76}{\scriptsize $\pm 0.77$} \\

Small & Random
& 78.78{\scriptsize $\pm 0.20$}
& 76.78{\scriptsize $\pm 0.20$}
& 74.31{\scriptsize $\pm 0.31$}
& 68.09{\scriptsize $\pm 0.20$}
& 63.40{\scriptsize $\pm 0.32$}
& 53.02{\scriptsize $\pm 0.62$} \\

Small & Large
& 79.12{\scriptsize $\pm 0.28$}
& 77.27{\scriptsize $\pm 0.12$}
& 74.31{\scriptsize $\pm 0.30$}
& 68.75{\scriptsize $\pm 0.27$}
& 64.24{\scriptsize $\pm 0.25$}
& 53.74{\scriptsize $\pm 0.56$} \\

Large & Small
& 79.11{\scriptsize $\pm 0.17$}
& 76.78{\scriptsize $\pm 0.24$}
& 73.39{\scriptsize $\pm 0.18$}
& 66.22{\scriptsize $\pm 0.33$}
& 58.62{\scriptsize $\pm 0.46$}
& 39.94{\scriptsize $\pm 0.55$} \\

Large & Large
& 79.12{\scriptsize $\pm 0.28$}
& 76.62{\scriptsize $\pm 0.20$}
& 73.25{\scriptsize $\pm 0.45$}
& 64.09{\scriptsize $\pm 0.25$}
& 55.45{\scriptsize $\pm 0.63$}
& 35.96{\scriptsize $\pm 1.14$} \\

\bottomrule
\end{tabular}

\end{table}

\section{Extensions and Generalization}
\label{sec:app_extension}

\subsection{Generalization Across Architectures and Datasets}{
\label{sec:app_broad_exp}

We further evaluate DVA in the single-teacher setting across different
architectures and dataset scales. While the main text reports gains over
random selection for clarity, \Cref{tab:app_cifar100_vit_main,tab:app_tinyimagenet_resnet_main}
provide the corresponding absolute KD accuracies on CIFAR-100 with ViT
and on Tiny-ImageNet with ResNet, respectively.

Across both settings, DVA maintains strong performance over a wide range of pruning rates without substantial architecture- or dataset-specific retuning. On CIFAR-100 with ViT, DVA achieves the best accuracy among training-dynamics-free methods at five of the six pruning rates, with its advantage becoming particularly clear under aggressive pruning. A similar trend is observed on Tiny-ImageNet,  where DVA outperforms random selection at every pruning rate and achieves its largest gains under aggressive pruning. These results indicate that the proposed selection criteria generalize beyond the ResNet/CIFAR-100 setting used for our primary controlled experiments.

\begin{table}[H]
\centering
\caption{
KD accuracy (\%) on CIFAR-100 with a ViT-Ti student and a fine-tuned ViT-S teacher.
Results are averaged over 3 seeds and reported as mean $\pm$ standard deviation.
Best/second-best results within each group are in \textbf{bold}/\underline{underline}.
}
\label{tab:app_cifar100_vit_main}

\small
\renewcommand{\arraystretch}{1.05}
\setlength{\tabcolsep}{5pt}

\begin{tabular}{l cccccc}
\toprule
\multirow{2}{*}{Method} &
\multicolumn{6}{c}{Drop Rate} \\
\cmidrule(lr){2-7}
& 0.1 & 0.3 & 0.5 & 0.7 & 0.8 & 0.9 \\
\midrule

\multicolumn{7}{l}{\textit{Training-Dynamics-Based}} \\

D$^2$
& {\ul 73.56}{\scriptsize $\pm$0.24}
& \textbf{71.14}{\scriptsize $\pm$1.05}
& \textbf{66.60}{\scriptsize $\pm$0.88}
& \textbf{60.24}{\scriptsize $\pm$0.40}
& \textbf{55.30}{\scriptsize $\pm$0.12}
& \textbf{44.75}{\scriptsize $\pm$1.11} \\

DUAL
& \textbf{73.97}{\scriptsize $\pm$0.57}
& {\ul 70.78}{\scriptsize $\pm$0.39}
& {\ul 65.79}{\scriptsize $\pm$0.20}
& {\ul 59.09}{\scriptsize $\pm$0.37}
& {\ul 53.42}{\scriptsize $\pm$0.18}
& {\ul 44.00}{\scriptsize $\pm$0.53} \\

\midrule

\multicolumn{7}{l}{\textit{Training-Dynamics-Free}} \\

Random
& {\ul 73.79}{\scriptsize $\pm$0.72}
& 70.95{\scriptsize $\pm$0.23}
& 66.75{\scriptsize $\pm$0.40}
& {\ul 60.11}{\scriptsize $\pm$0.41}
& {\ul 54.28}{\scriptsize $\pm$0.67}
& {\ul 41.66}{\scriptsize $\pm$0.87} \\

MDSLR
& 73.71{\scriptsize $\pm$0.41}
& {\ul 71.16}{\scriptsize $\pm$0.12}
& {\ul 66.98}{\scriptsize $\pm$0.16}
& 59.90{\scriptsize $\pm$0.48}
& 53.29{\scriptsize $\pm$0.79}
& 40.74{\scriptsize $\pm$0.53} \\

IF-Beta
& \textbf{74.10}{\scriptsize $\pm$1.07}
& 70.12{\scriptsize $\pm$1.40}
& 60.67{\scriptsize $\pm$0.32}
& 53.69{\scriptsize $\pm$0.59}
& 50.91{\scriptsize $\pm$0.54}
& 40.11{\scriptsize $\pm$0.79} \\

\rowcolor{gray!12}
DVA (Ours)
& {\ul 73.79}{\scriptsize $\pm$0.14}
& \textbf{71.24}{\scriptsize $\pm$0.87}
& \textbf{67.28}{\scriptsize $\pm$0.65}
& \textbf{61.50}{\scriptsize $\pm$0.77}
& \textbf{55.41}{\scriptsize $\pm$0.24}
& \textbf{45.82}{\scriptsize $\pm$0.62} \\

\bottomrule
\end{tabular}
\end{table}

\begin{table}[H]
\centering
\caption{
KD accuracy (\%) on Tiny-ImageNet using the ResNet-34 teacher.
Results are averaged over 3 seeds and reported as mean $\pm$ standard deviation.
Best/second-best results within each group are in \textbf{bold}/\underline{underline}.
}
\label{tab:app_tinyimagenet_resnet_main}

\small
\renewcommand{\arraystretch}{1.05}
\setlength{\tabcolsep}{5pt}

\begin{tabular}{l cccccc}
\toprule
\multirow{2}{*}{Method} &
\multicolumn{6}{c}{Drop Rate} \\
\cmidrule(lr){2-7}
& 0.1 & 0.3 & 0.5 & 0.7 & 0.8 & 0.9 \\
\midrule

\multicolumn{7}{l}{\textit{Training-Dynamics-Based}} \\

D$^2$
& {\ul 65.40}{\scriptsize $\pm$0.09}
& {\ul 63.64}{\scriptsize $\pm$0.35}
& {\ul 62.38}{\scriptsize $\pm$0.21}
& {\ul 58.43}{\scriptsize $\pm$0.09}
& {\ul 54.34}{\scriptsize $\pm$0.22}
& \textbf{47.37}{\scriptsize $\pm$0.31} \\

DUAL
& \textbf{66.32}{\scriptsize $\pm$0.25}
& \textbf{65.26}{\scriptsize $\pm$0.16}
& \textbf{63.49}{\scriptsize $\pm$0.31}
& \textbf{59.49}{\scriptsize $\pm$0.42}
& \textbf{55.32}{\scriptsize $\pm$0.24}
& {\ul 46.30}{\scriptsize $\pm$0.17} \\

\midrule

\multicolumn{7}{l}{\textit{Training-Dynamics-Free}} \\

Random
& 66.20{\scriptsize $\pm$0.47}
& 64.45{\scriptsize $\pm$0.40}
& 62.23{\scriptsize $\pm$0.33}
& 57.45{\scriptsize $\pm$0.24}
& 54.11{\scriptsize $\pm$0.36}
& 45.03{\scriptsize $\pm$0.64} \\

MDSLR
& \textbf{66.38}{\scriptsize $\pm$0.22}
& {\ul 64.74}{\scriptsize $\pm$0.26}
& 62.11{\scriptsize $\pm$0.19}
& 57.97{\scriptsize $\pm$0.46}
& 54.05{\scriptsize $\pm$0.21}
& 44.86{\scriptsize $\pm$0.36} \\

IF-Beta
& 66.27{\scriptsize $\pm$0.10}
& 64.02{\scriptsize $\pm$0.30}
& \textbf{63.62}{\scriptsize $\pm$0.21}
& \textbf{59.81}{\scriptsize $\pm$0.12}
& {\ul 55.85}{\scriptsize $\pm$0.38}
& {\ul 47.38}{\scriptsize $\pm$0.11} \\

\rowcolor{gray!12}
DVA (Ours)
& {\ul 66.32}{\scriptsize $\pm$0.15}
& \textbf{65.24}{\scriptsize $\pm$0.27}
& {\ul 63.31}{\scriptsize $\pm$0.10}
& {\ul 59.65}{\scriptsize $\pm$0.31}
& \textbf{56.26}{\scriptsize $\pm$0.31}
& \textbf{48.88}{\scriptsize $\pm$0.45} \\

\bottomrule
\end{tabular}
\end{table}

}

\subsection{Multi-Teacher KD}
\label{sec:app_mtkd}

In this section, we further evaluate data selection under a simple scheduled multi-teacher KD setting. A natural question raised by our previous results is whether the low-data degradation can be addressed simply by replacing the large teacher with a smaller one, or whether data selection can provide additional gains even when the teacher capacity is adjusted to the data budget. Motivated by the budget-dependent teacher preference observed in \Cref{fig:overview_a} and further examined in \Cref{sec:app_teacher_sweep_kd}, we therefore consider a simple schedule that gradually shifts supervision from a large teacher to a small teacher as the data budget decreases.

Let $\rho$ denote the drop rate. At $\rho=0.1$, the student is supervised only by the large teacher, while at $\rho=0.9$, it is supervised only by the small teacher. For intermediate drop rates, we linearly interpolate the two KD losses, assigning progressively more weight to the small teacher as the data budget decreases. Specifically, the large teacher weight is
$\lambda_\rho = \frac{0.9-\rho}{0.8}$, and the small teacher weight is $1-\lambda_\rho$. The overall objective is

\begin{equation}
   \mathcal{L}
   =
   0.5\mathcal{L}_{\mathrm{CE}}
   +
   0.5\left[
   \lambda_\rho \mathcal{L}_{\mathrm{KD}}^{\mathrm{large}}
   +
   (1-\lambda_\rho)\mathcal{L}_{\mathrm{KD}}^{\mathrm{small}}
   \right].
\end{equation}

This schedule is intentionally simple: it directly reflects the observed shift in teacher preference while recovering single teacher supervision at the two endpoints.

As shown in \Cref{tab:cifar100_ensemble_sched,tab:cifar100_vit_ensemble_sched,tab:tinyimagenet_resnet_ensemble_sched}, the effect of scheduled multi-teacher supervision itself varies across settings. Its benefit is most evident in the low-data regime, where the schedule places greater weight on the smaller teacher. In several settings, this substantially mitigates the performance degradation under moderate to aggressive pruning, consistent with the budget-dependent teacher preference observed in our earlier analysis.

More importantly, data selection continues to provide additional gains under scheduled multi-teacher supervision. On CIFAR-100 with the ResNet student (\Cref{tab:cifar100_ensemble_sched}), our method consistently improves over random selection across all drop rates and achieves the best training-dynamics-free performance in five of the six drop rates. At a drop rate of $0.9$, where supervision comes entirely from the small teacher and random selection already reaches $62.49\%$, DVA performs comparably ($62.58\%$), suggesting that the small teacher already provides highly effective supervision under this severe data constraint. The benefit becomes more pronounced for the ViT student in the low-data regime (\Cref{tab:cifar100_vit_ensemble_sched}): compared with random selection, our method improves KD accuracy by $+1.15$ and $+2.55$ pp at drop rates of $0.8$ and $0.9$, respectively, while achieving the best training-dynamics-free performance throughout this range. A similar trend appears on Tiny-ImageNet (\Cref{tab:tinyimagenet_resnet_ensemble_sched}), where our method improves over random selection by $+1.24$ and $+1.79$ pp at drop rates of $0.8$ and $0.9$, respectively, and remains the best or second-best training-dynamics-free method across all budgets. These results show that adapting teacher supervision to the data budget does not make data selection redundant; selecting an effective subset can provide additional gains even when the contribution of the smaller teacher is explicitly increased in the low-data regime.

\begin{table}[H]
\centering
\caption{
KD accuracy (\%) on CIFAR-100 with a ResNet-18 student under scheduled multi-teacher supervision. Results are averaged over 10 seeds and reported as mean $\pm$ standard deviation. Best/second-best results are in \textbf{bold}/\underline{underline}.
}
\label{tab:cifar100_ensemble_sched}

\footnotesize
\renewcommand{\arraystretch}{1.10}
\setlength{\tabcolsep}{5pt}

\begin{tabular}{l cccccc}
\toprule
\multirow{2}{*}{Method} &
\multicolumn{6}{c}{Drop Rate} \\
\cmidrule(lr){2-7}
& 0.1 & 0.3 & 0.5 & 0.7 & 0.8 & 0.9 \\
\midrule

\multicolumn{7}{l}{\textit{Single-Teacher KD (Random)}} \vspace{1pt} \\

Large T.
& 78.82{\scriptsize$\pm$0.25}
& 76.69{\scriptsize$\pm$0.17}
& 73.16{\scriptsize$\pm$0.20}
& 67.07{\scriptsize$\pm$0.49}
& 61.35{\scriptsize$\pm$0.43}
& 46.93{\scriptsize$\pm$0.80} \\

Small T.
& 75.33{\scriptsize$\pm$0.19}
& 74.64{\scriptsize$\pm$0.13}
& 73.54{\scriptsize$\pm$0.11}
& 71.24{\scriptsize$\pm$0.23}
& 69.09{\scriptsize$\pm$0.16}
& 62.49{\scriptsize$\pm$0.41} \\

\midrule

\multicolumn{7}{l}{\textit{Training-Dynamics-Based}} \vspace{1pt} \\

Forgetting
& \textbf{79.35}{\scriptsize$\pm$0.27}
& \textbf{78.48}{\scriptsize$\pm$0.22}
& {\ul 75.71}{\scriptsize$\pm$0.25}
& 70.16{\scriptsize$\pm$0.23}
& 65.99{\scriptsize$\pm$0.08}
& 57.37{\scriptsize$\pm$0.33} \\

EL2N
& 79.06{\scriptsize$\pm$0.28}
& 77.80{\scriptsize$\pm$0.22}
& 73.84{\scriptsize$\pm$0.31}
& 65.83{\scriptsize$\pm$0.52}
& 60.25{\scriptsize$\pm$0.21}
& 46.96{\scriptsize$\pm$0.86} \\

CCS
& 79.13{\scriptsize$\pm$0.23}
& 78.40{\scriptsize$\pm$0.26}
& 75.63{\scriptsize$\pm$0.08}
& 72.41{\scriptsize$\pm$0.24}
& 69.21{\scriptsize$\pm$0.20}
& {\ul 62.88}{\scriptsize$\pm$0.51} \\

D$^2$
& 79.02{\scriptsize$\pm$0.27}
& 78.00{\scriptsize$\pm$0.20}
& 75.68{\scriptsize$\pm$0.15}
& \textbf{72.81}{\scriptsize$\pm$0.18}
& \textbf{70.00}{\scriptsize$\pm$0.14}
& \textbf{63.67}{\scriptsize$\pm$0.27} \\

DUAL
& {\ul 79.29}{\scriptsize$\pm$0.28}
& {\ul 78.45}{\scriptsize$\pm$0.24}
& \textbf{76.19}{\scriptsize$\pm$0.14}
& {\ul 72.43}{\scriptsize$\pm$0.15}
& {\ul 69.55}{\scriptsize$\pm$0.09}
& 62.79{\scriptsize$\pm$0.23} \\

\midrule

\multicolumn{7}{l}{\textit{Training-Dynamics-Free}} \vspace{1pt} \\

Random
& 78.82{\scriptsize$\pm$0.25}
& 77.45{\scriptsize$\pm$0.24}
& {\ul 75.20}{\scriptsize$\pm$0.28}
& {\ul 71.74}{\scriptsize$\pm$0.26}
& 69.08{\scriptsize$\pm$0.38}
& 62.49{\scriptsize$\pm$0.41} \\

GraNd
& 78.64{\scriptsize$\pm$0.32}
& 76.10{\scriptsize$\pm$0.16}
& 72.87{\scriptsize$\pm$0.08}
& 68.52{\scriptsize$\pm$0.28}
& 64.53{\scriptsize$\pm$0.38}
& 52.61{\scriptsize$\pm$0.57} \\

MDSLR
& 78.58{\scriptsize$\pm$0.25}
& 77.07{\scriptsize$\pm$0.14}
& 74.78{\scriptsize$\pm$0.21}
& 71.48{\scriptsize$\pm$0.26}
& 69.26{\scriptsize$\pm$0.23}
& 62.33{\scriptsize$\pm$0.28} \\

IF-Beta
& {\ul 79.05}{\scriptsize$\pm$0.21}
& {\ul 77.78}{\scriptsize$\pm$0.19}
& 74.29{\scriptsize$\pm$0.25}
& 71.03{\scriptsize$\pm$0.34}
& \textbf{69.43}{\scriptsize$\pm$0.40}
& {\ul 62.57}{\scriptsize$\pm$0.66} \\

\rowcolor{gray!12}
DVA (Ours)
& \textbf{79.13}{\scriptsize$\pm$0.17}
& \textbf{77.96}{\scriptsize$\pm$0.22}
& \textbf{75.79}{\scriptsize$\pm$0.18}
& \textbf{72.06}{\scriptsize$\pm$0.20}
& {\ul 69.34}{\scriptsize$\pm$0.17}
& \textbf{62.58}{\scriptsize$\pm$0.72} \\

\bottomrule
\end{tabular}
\end{table}

\begin{table}[H]
\centering
\caption{
KD accuracy (\%) on CIFAR-100 with a ViT-Ti student under
scheduled multi-teacher supervision.
Results are averaged over 3 seeds and reported as mean $\pm$ standard deviation.
Best/second-best results within each group are in \textbf{bold}/\underline{underline}.
}
\label{tab:cifar100_vit_ensemble_sched}

\small
\renewcommand{\arraystretch}{1.05}
\setlength{\tabcolsep}{5pt}

\begin{tabular}{l cccccc}
\toprule
\multirow{2}{*}{Method} &
\multicolumn{6}{c}{Drop Rate} \\
\cmidrule(lr){2-7}
& 0.1 & 0.3 & 0.5 & 0.7 & 0.8 & 0.9 \\
\midrule

\multicolumn{7}{l}{\textit{Training-Dynamics-Based}} \\

D$^2$
& {\ul 73.56}{\scriptsize $\pm$0.24}
& \textbf{70.87}{\scriptsize $\pm$0.74}
& \textbf{65.97}{\scriptsize $\pm$0.47}
& \textbf{59.50}{\scriptsize $\pm$0.13}
& \textbf{54.89}{\scriptsize $\pm$0.20}
& \textbf{45.96}{\scriptsize $\pm$0.31} \\

DUAL
& \textbf{73.97}{\scriptsize $\pm$0.57}
& {\ul 70.58}{\scriptsize $\pm$0.23}
& {\ul 65.72}{\scriptsize $\pm$0.61}
& {\ul 58.26}{\scriptsize $\pm$0.13}
& {\ul 52.95}{\scriptsize $\pm$0.09}
& {\ul 44.15}{\scriptsize $\pm$0.30} \\

\midrule

\multicolumn{7}{l}{\textit{Training-Dynamics-Free}} \\

Random
& {\ul 73.79}{\scriptsize $\pm$0.72}
& {\ul 70.99}{\scriptsize $\pm$0.63}
& 65.97{\scriptsize $\pm$0.27}
& 58.90{\scriptsize $\pm$0.32}
& {\ul 54.01}{\scriptsize $\pm$0.47}
& 44.05{\scriptsize $\pm$0.44} \\

MDSLR
& 73.71{\scriptsize $\pm$0.41}
& \textbf{71.00}{\scriptsize $\pm$0.66}
& {\ul 66.19}{\scriptsize $\pm$0.50}
& {\ul 59.71}{\scriptsize $\pm$0.23}
& {\ul 54.01}{\scriptsize $\pm$0.36}
& {\ul 44.19}{\scriptsize $\pm$0.23} \\

IF-Beta
& \textbf{74.10}{\scriptsize $\pm$1.07}
& 70.41{\scriptsize $\pm$0.43}
& 59.69{\scriptsize $\pm$0.33}
& 54.42{\scriptsize $\pm$0.49}
& 52.08{\scriptsize $\pm$0.58}
& 42.51{\scriptsize $\pm$0.52} \\

\rowcolor{gray!12}
DVA (Ours)
& {\ul 73.79}{\scriptsize $\pm$0.14}
& 70.58{\scriptsize $\pm$0.70}
& \textbf{66.53}{\scriptsize $\pm$0.71}
& \textbf{60.44}{\scriptsize $\pm$0.44}
& \textbf{55.16}{\scriptsize $\pm$0.24}
& \textbf{46.60}{\scriptsize $\pm$0.45} \\

\bottomrule
\end{tabular}
\end{table}

\begin{table}[H]
\centering
\caption{
KD accuracy (\%) on Tiny-ImageNet with a ResNet student under scheduled multi-teacher supervision.
Results are averaged over 3 seeds and reported as mean $\pm$ standard deviation.
Best/second-best results within each group are in \textbf{bold}/\underline{underline}.
}
\label{tab:tinyimagenet_resnet_ensemble_sched}

\small
\renewcommand{\arraystretch}{1.05}
\setlength{\tabcolsep}{5pt}

\begin{tabular}{l cccccc}
\toprule
\multirow{2}{*}{Method} &
\multicolumn{6}{c}{Drop Rate} \\
\cmidrule(lr){2-7}
& 0.1 & 0.3 & 0.5 & 0.7 & 0.8 & 0.9 \\
\midrule

\multicolumn{7}{l}{\textit{Training-Dynamics-Based}} \\

D$^2$
& {\ul 65.40}{\scriptsize $\pm$0.09}
& {\ul 64.15}{\scriptsize $\pm$0.20}
& {\ul 63.64}{\scriptsize $\pm$0.27}
& {\ul 60.75}{\scriptsize $\pm$0.25}
& {\ul 57.92}{\scriptsize $\pm$0.23}
& \textbf{52.87}{\scriptsize $\pm$0.23} \\

DUAL
& \textbf{66.32}{\scriptsize $\pm$0.25}
& \textbf{65.84}{\scriptsize $\pm$0.19}
& \textbf{64.68}{\scriptsize $\pm$0.36}
& \textbf{61.34}{\scriptsize $\pm$0.03}
& \textbf{58.03}{\scriptsize $\pm$0.30}
& {\ul 52.69}{\scriptsize $\pm$0.09} \\

\midrule

\multicolumn{7}{l}{\textit{Training-Dynamics-Free}} \\

Random
& 66.20{\scriptsize $\pm$0.47}
& 65.25{\scriptsize $\pm$0.08}
& 63.71{\scriptsize $\pm$0.18}
& 60.70{\scriptsize $\pm$0.46}
& 57.85{\scriptsize $\pm$0.21}
& 52.63{\scriptsize $\pm$0.09} \\

MDSLR
& \textbf{66.38}{\scriptsize $\pm$0.22}
& {\ul 65.72}{\scriptsize $\pm$0.39}
& 64.25{\scriptsize $\pm$0.17}
& 60.84{\scriptsize $\pm$0.50}
& 58.28{\scriptsize $\pm$0.40}
& 52.31{\scriptsize $\pm$0.31} \\

IF-Beta
& 66.27{\scriptsize $\pm$0.10}
& 64.57{\scriptsize $\pm$0.15}
& \textbf{64.69}{\scriptsize $\pm$0.24}
& \textbf{61.64}{\scriptsize $\pm$0.19}
& {\ul 58.82}{\scriptsize $\pm$0.45}
& {\ul 54.00}{\scriptsize $\pm$0.73} \\

\rowcolor{gray!12}
DVA (Ours)
& {\ul 66.32}{\scriptsize $\pm$0.15}
& \textbf{66.30}{\scriptsize $\pm$0.13}
& {\ul 64.52}{\scriptsize $\pm$0.27}
& {\ul 61.32}{\scriptsize $\pm$0.22}
& \textbf{59.09}{\scriptsize $\pm$0.14}
& \textbf{54.42}{\scriptsize $\pm$0.26} \\

\bottomrule
\end{tabular}
\end{table}

\subsection{Cross-Architecture Transfer}
\label{sec:app_cross_kd}

We further evaluate whether the selected subsets generalize across student architectures. We follow the same CIFAR-100 setup as in the ResNet experiments, but replace the student with VGG-11~\citep{simonyan2014very} and MobileNetV2~\citep{sandler2018mobilenetv2}. Importantly, we reuse the subsets selected under the original ResNet-based selection setup without rerunning data selection for each student architecture; thus, only the student architecture is changed. This setting allows us to examine whether the benefit of our selection method is tied to a particular student or transfers across architectures.

As shown in \Cref{tab:cifar100_vgg11,tab:cifar100_mobilenetv2}, our method remains consistently effective across both students. For VGG-11 (\Cref{tab:cifar100_vgg11}), it achieves the best performance among training-dynamics-free methods at all six pruning rates, improving over random selection by 0.21, 0.76, 1.48, 3.02, 3.65, and 5.88 percentage points as the drop rate increases from 0.1 to 0.9. The advantage becomes particularly pronounced under aggressive pruning, reaching 49.28\% accuracy at a 0.9 drop rate compared with 43.40\% for random selection and 45.53\% for the strongest competing training-dynamics-free baseline. A similar trend is observed with MobileNetV2 (\Cref{tab:cifar100_mobilenetv2}): our method is best among training-dynamics-free approaches at four of the six pruning rates and second-best at the remaining two, while improving over random selection by 5.34 points at a 0.9 drop rate (29.73\% vs. 24.39\%). These results indicate that the selected subsets are not specialized to the student architecture used in the original setup, and that the benefit of our training-dynamics-free selection transfers to substantially different student architectures.

\begin{table}[H]
\centering
\caption{
KD accuracy (\%) on CIFAR-100 with a VGG-11 student.
Data pruning is performed using the ResNet-based selection setup.
Results are averaged over 3 seeds and reported as mean $\pm$ standard deviation.
Best/second-best results within each group are in \textbf{bold}/\underline{underline}.
}
\label{tab:cifar100_vgg11}

\small
\renewcommand{\arraystretch}{1.08}
\setlength{\tabcolsep}{5pt}

\begin{tabular}{lcccccc}
\toprule
\multirow{2}{*}{Method}
& \multicolumn{6}{c}{Drop Rate} \\
\cmidrule(lr){2-7}
& 0.1 & 0.3 & 0.5 & 0.7 & 0.8 & 0.9 \\
\midrule

\multicolumn{7}{l}{\textit{Training-Dynamics-Based}} \\

D$^2$
& {\ul 73.09}{\scriptsize $\pm$0.26}
& {\ul 71.58}{\scriptsize $\pm$0.27}
& {\ul 68.80}{\scriptsize $\pm$0.40}
& {\ul 62.70}{\scriptsize $\pm$0.24}
& {\ul 59.49}{\scriptsize $\pm$0.09}
& {\ul 49.34}{\scriptsize $\pm$0.66} \\

DUAL
& \textbf{73.40}{\scriptsize $\pm$0.25}
& \textbf{71.84}{\scriptsize $\pm$0.21}
& \textbf{68.87}{\scriptsize $\pm$0.15}
& \textbf{64.68}{\scriptsize $\pm$0.09}
& \textbf{59.98}{\scriptsize $\pm$0.30}
& \textbf{50.80}{\scriptsize $\pm$0.07} \\

\midrule
\multicolumn{7}{l}{\textit{Training-Dynamics-Free}} \\

Random
& {\ul 73.00}{\scriptsize $\pm$0.07}
& 70.22{\scriptsize $\pm$0.12}
& {\ul 66.10}{\scriptsize $\pm$0.44}
& 59.89{\scriptsize $\pm$0.11}
& 54.90{\scriptsize $\pm$0.37}
& 43.40{\scriptsize $\pm$1.20} \\

MDSLR
& 72.65{\scriptsize $\pm$0.34}
& {\ul 70.24}{\scriptsize $\pm$0.13}
& 65.91{\scriptsize $\pm$0.16}
& {\ul 60.29}{\scriptsize $\pm$0.16}
& 55.84{\scriptsize $\pm$0.19}
& {\ul 45.53}{\scriptsize $\pm$0.40} \\

IF-Beta
& 72.89{\scriptsize $\pm$0.16}
& 69.21{\scriptsize $\pm$0.09}
& 61.73{\scriptsize $\pm$0.08}
& 56.49{\scriptsize $\pm$0.67}
& {\ul 56.56}{\scriptsize $\pm$0.16}
& 44.79{\scriptsize $\pm$0.18} \\

\rowcolor{gray!12}
DVA (Ours)
& \textbf{73.21}{\scriptsize $\pm$0.26}
& \textbf{70.98}{\scriptsize $\pm$0.09}
& \textbf{67.58}{\scriptsize $\pm$0.34}
& \textbf{62.91}{\scriptsize $\pm$0.45}
& \textbf{58.55}{\scriptsize $\pm$0.25}
& \textbf{49.28}{\scriptsize $\pm$0.12} \\

\bottomrule
\end{tabular}
\end{table}
\begin{table}[H]
\centering
\caption{
KD accuracy (\%) on CIFAR-100 with a MobileNetV2 student.
Data pruning is performed using the ResNet-based selection setup.
Results are averaged over 3 seeds and reported as mean $\pm$ standard deviation.
Best/second-best results within each group are in \textbf{bold}/\underline{underline}.
}
\label{tab:cifar100_mobilenetv2}

\small
\renewcommand{\arraystretch}{1.08}
\setlength{\tabcolsep}{5pt}

\begin{tabular}{lcccccc}
\toprule
\multirow{2}{*}{Method}
& \multicolumn{6}{c}{Drop Rate} \\
\cmidrule(lr){2-7}
& 0.1 & 0.3 & 0.5 & 0.7 & 0.8 & 0.9 \\
\midrule

\multicolumn{7}{l}{\textit{Training-Dynamics-Based}} \\

D$^2$
& {\ul 65.20}{\scriptsize $\pm$0.99}
& \textbf{64.11}{\scriptsize $\pm$0.35}
& \textbf{59.12}{\scriptsize $\pm$0.14}
& {\ul 47.39}{\scriptsize $\pm$0.80}
& {\ul 40.93}{\scriptsize $\pm$1.44}
& {\ul 30.52}{\scriptsize $\pm$0.66} \\

DUAL
& \textbf{66.15}{\scriptsize $\pm$0.79}
& {\ul 63.93}{\scriptsize $\pm$0.28}
& {\ul 58.69}{\scriptsize $\pm$0.37}
& \textbf{49.74}{\scriptsize $\pm$0.99}
& \textbf{42.64}{\scriptsize $\pm$0.78}
& \textbf{31.89}{\scriptsize $\pm$0.72} \\

\midrule
\multicolumn{7}{l}{\textit{Training-Dynamics-Free}} \\

Random
& 65.19{\scriptsize $\pm$0.81}
& 61.63{\scriptsize $\pm$0.52}
& 55.11{\scriptsize $\pm$0.09}
& {\ul 45.97}{\scriptsize $\pm$0.25}
& 35.92{\scriptsize $\pm$0.19}
& 24.39{\scriptsize $\pm$2.00} \\

MDSLR
& 65.34{\scriptsize $\pm$0.14}
& {\ul 62.02}{\scriptsize $\pm$0.17}
& \textbf{56.93}{\scriptsize $\pm$1.06}
& 45.79{\scriptsize $\pm$0.67}
& 37.53{\scriptsize $\pm$0.38}
& 25.69{\scriptsize $\pm$1.24} \\

IF-Beta
& \textbf{65.64}{\scriptsize $\pm$0.04}
& 61.03{\scriptsize $\pm$0.54}
& 51.98{\scriptsize $\pm$0.92}
& 40.27{\scriptsize $\pm$0.34}
& {\ul 39.72}{\scriptsize $\pm$0.51}
& {\ul 27.06}{\scriptsize $\pm$0.26} \\

\rowcolor{gray!12}
DVA (Ours)
& {\ul 65.49}{\scriptsize $\pm$0.34}
& \textbf{62.67}{\scriptsize $\pm$0.77}
& {\ul 56.86}{\scriptsize $\pm$1.05}
& \textbf{47.86}{\scriptsize $\pm$0.53}
& \textbf{40.39}{\scriptsize $\pm$0.43}
& \textbf{29.73}{\scriptsize $\pm$0.71} \\

\bottomrule
\end{tabular}
\end{table}

\section{Computational Cost and Scalability}

To evaluate the practical scalability of DVA, we measure the computational cost of data selection on ImageNet at a 0.9 drop rate. For this experiment, we directly use an off-the-shelf pretrained model as the proxy, without any additional proxy training or adaptation. We compare DVA with IF-Beta, which requires validation set adaptation, gradient- and Hessian-based influence estimation, and surrogate student optimization during data selection. DVA requires only 2.81 PFLOPs, compared with 71.41 PFLOPs for IF-Beta, corresponding to a $25.4\times$ reduction in selection cost. This reduction is also reflected in wall-clock time, decreasing from 5{,}370\,s to 94\,s ($57.1\times$). These differences are particularly relevant for large-scale settings, where the additional optimization and gradient-based computations required by IF-Beta can make data selection increasingly expensive. In contrast, DVA performs selection solely from fixed outputs of the pretrained proxy and requires neither gradient computation nor additional model optimization, making it substantially easier to scale to large datasets and models.

\begin{table}[H]
\centering
\caption{
Computational cost of data selection on ImageNet at a 0.9 drop rate.
}
\label{tab:imagenet_selection_cost}

\small
\renewcommand{\arraystretch}{1.08}
\setlength{\tabcolsep}{10pt}

\begin{tabular}{lcc}
\toprule
Method & Selection Cost (PFLOPs) & Reduction \\
\midrule

IF-Beta
& 71.41
& - \\

DVA (Ours)
& \textbf{2.81}
& \textbf{$25.4\times$} \\

\bottomrule
\end{tabular}
\end{table}

\end{document}